\documentclass{article}

\usepackage{graphicx}
\usepackage{graphbox}
\usepackage{comment}
\usepackage{amsmath,amssymb} % define this before the line numbering.
\usepackage[dvipsnames]{xcolor}
\usepackage{multirow}
\usepackage{mathtools}
\usepackage{subcaption}
\usepackage{caption}
\DeclareCaptionFont{9pt}{\fontsize{9pt}{10pt}\selectfont}
\usepackage[affil-it]{authblk}
\usepackage{geometry}
\usepackage{titlesec}
\usepackage[utf8]{inputenc}
\usepackage[titletoc,toc,title]{appendix}

\usepackage{bm}
\usepackage{amsmath, amssymb}

\def\eg{{\em e.g.,~}}

\def\eg{{\em e.g.,~}}

\def\clevrchange{CLEVR-Change}
\def\std{ST\&D~}

\DeclareMathOperator*{\argmax}{arg\,max}

\newif\ifshowcomment
\showcommenttrue
\newcommand{\iaf}{I_{t}}
\newcommand{\ibe}{I_{t'}}
\usepackage{tikz}
\usetikzlibrary{positioning,fit,backgrounds,calc,intersections}

\makeatletter
\def\@maketitle{%
	\newpage
	\null
	\vskip 2em%
	\begin{center}%
		\let \footnote \thanks
		{\Large\bfseries \@title \par}%
		\vskip 1.5em%
		{\normalsize
			\lineskip .5em%
			\begin{tabular}[t]{c}%
				\@author
			\end{tabular}\par}%
		\vskip 1em%
		{\normalsize \@date}%
	\end{center}%
	\par
	\vskip 1.5em}
\makeatother

\title{Detection and Description of Change in Visual Streams} % Replace with your title

\author[1]{Davis Gilton}
\author[2]{Ruotian Luo}
\author[3]{Rebecca Willett}
\author[4]{Greg Shakhnarovich\thanks{All authors gratefully acknowledge support for this work by NSF Grant DMS-1925101 as well as AFOSR Grant FA9550-18-1-0166. In addition, DG and RW were supported by NSF Grant 1740707, and RL and GS were supported by the DARPA L2M program, award FA8750-18-2-0126.}}
\affil[1]{University of Wisconsin-Madison\\ gilton@wisc.edu}
\affil[2]{TTI-Chicago, rluo@ttic.edu}
\affil[3]{University of Chicago\\ willett@uchicago.edu}
\affil[4]{TTI-Chicago, gregory@ttic.edu}

\begin{document}

\maketitle

\begin{abstract}
	This paper presents a framework for the analysis of changes in
	visual streams: ordered sequences of images, possibly separated by
	significant time gaps. We propose
	a new approach to incorporating unlabeled data into training
	to generate natural language descriptions of change. We also develop a framework for estimating the time of
	change in visual stream. We use learned representations for
	change evidence and consistency of perceived change, and combine these
	in a regularized graph cut based change detector. Experimental
	evaluation on visual stream datasets, which we release as part of
	our contribution, shows that representation learning driven by natural
	language descriptions significantly improves change detection
	accuracy, compared to methods that do not rely on language. 
\end{abstract}
\section{Introduction}\label{sec:intro}
Visual streams depict changes in a scene over time, and this paper describes a learning-based approach to describing  those changes in a manner that (a) improves the detection changes beyond what is possible based solely on image analysis without change descriptions, (b) incorporates unlabled data to yield strong performance even with a limited number of labeled samples, and (c) is robust to nuisance changes such as lighting or perspective shifts. 

\begin{figure}
	\includegraphics[width=0.16\textwidth]{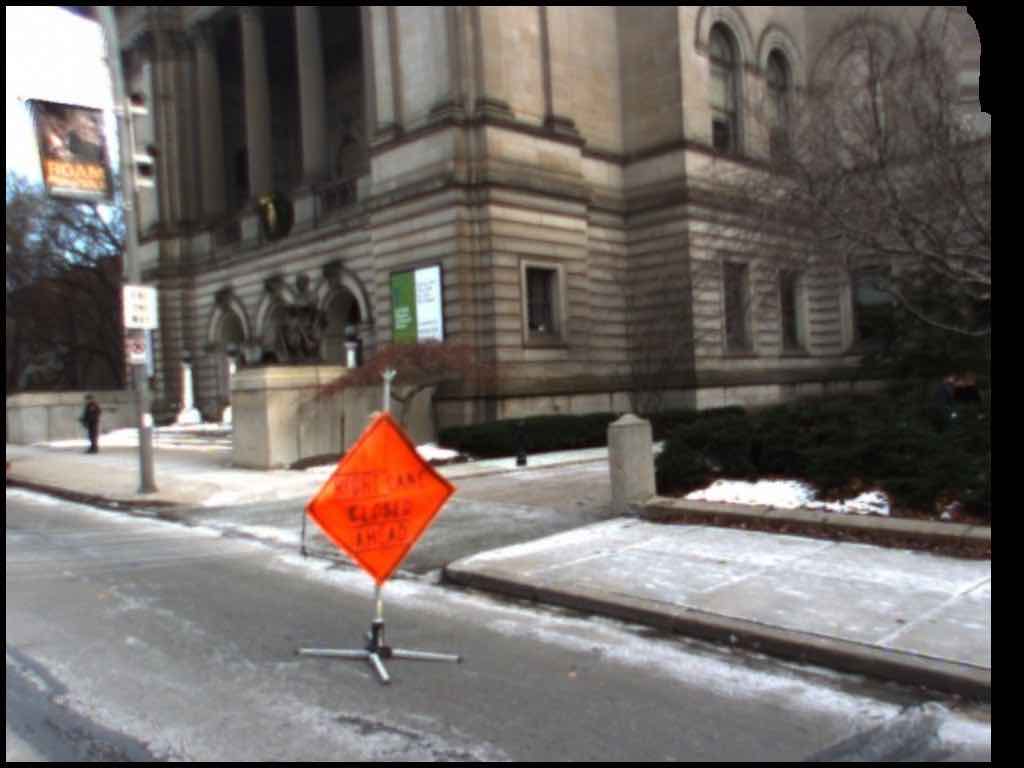}\hfill
	\includegraphics[width=0.16\textwidth]{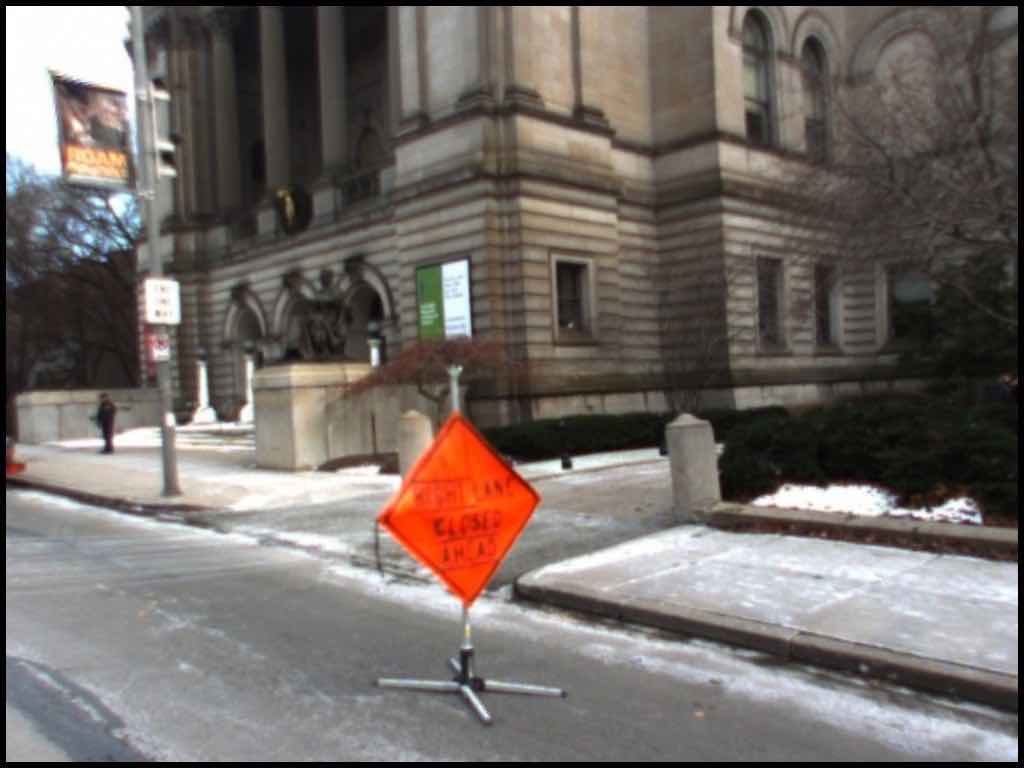}\hfill
	\includegraphics[width=0.16\textwidth]{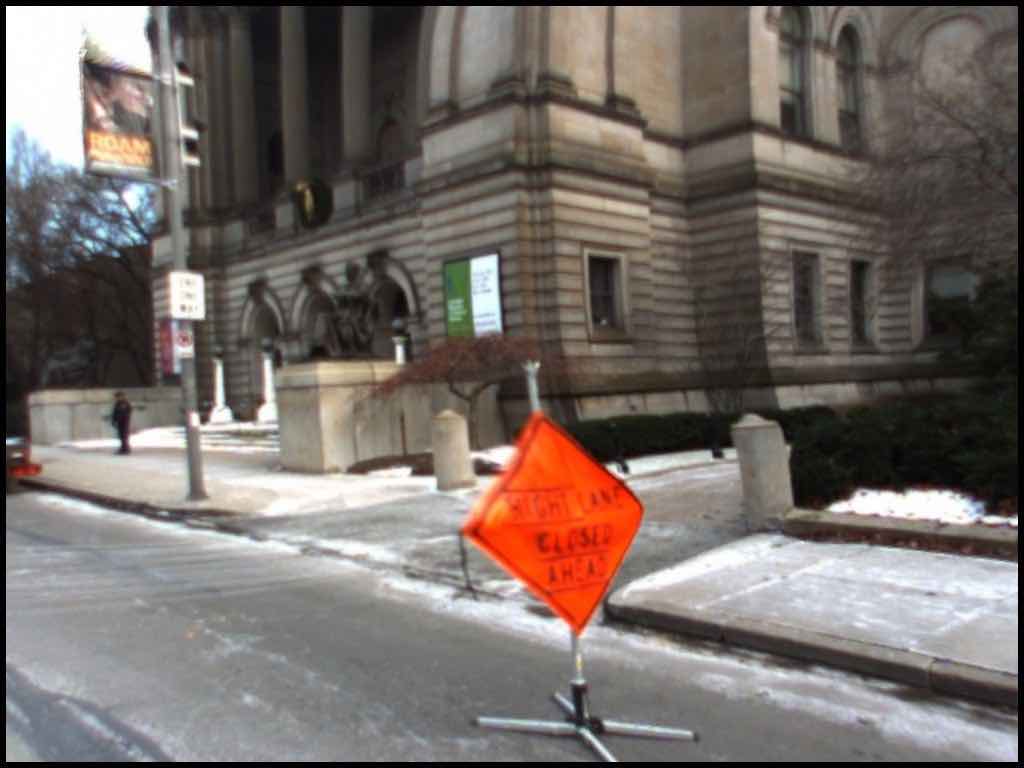}\hfill
	\includegraphics[width=0.16\textwidth]{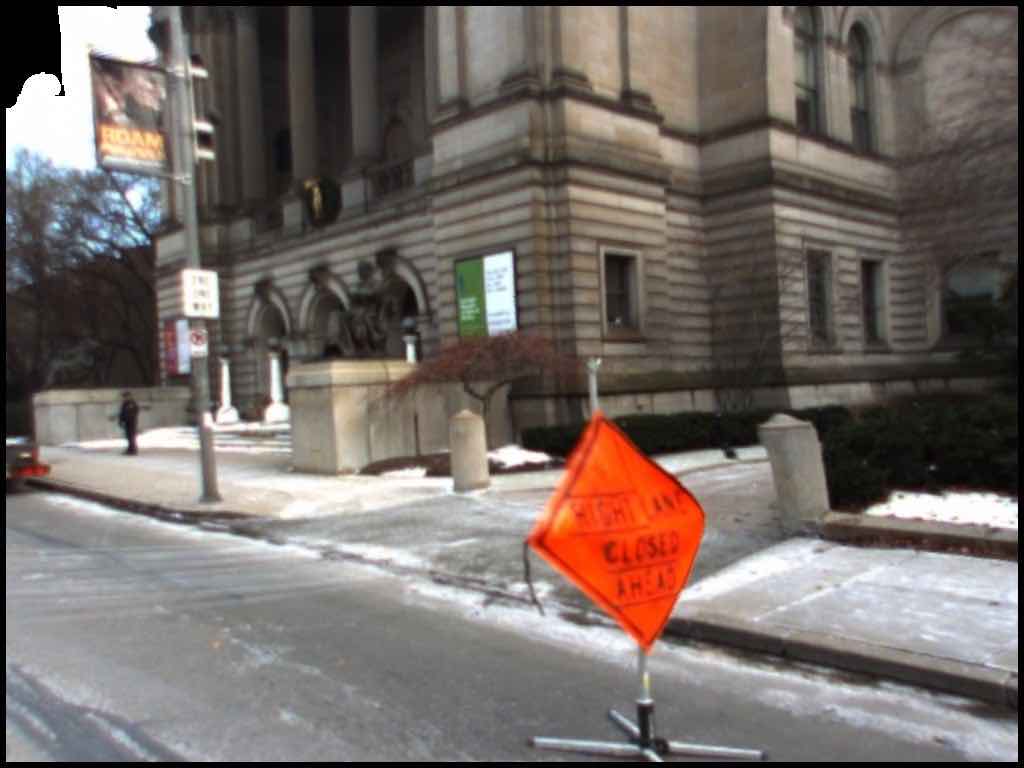}\hfill
	\includegraphics[height=5em,width=0.007\textwidth]{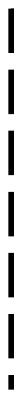}\hfill
	\includegraphics[width=0.16\textwidth]{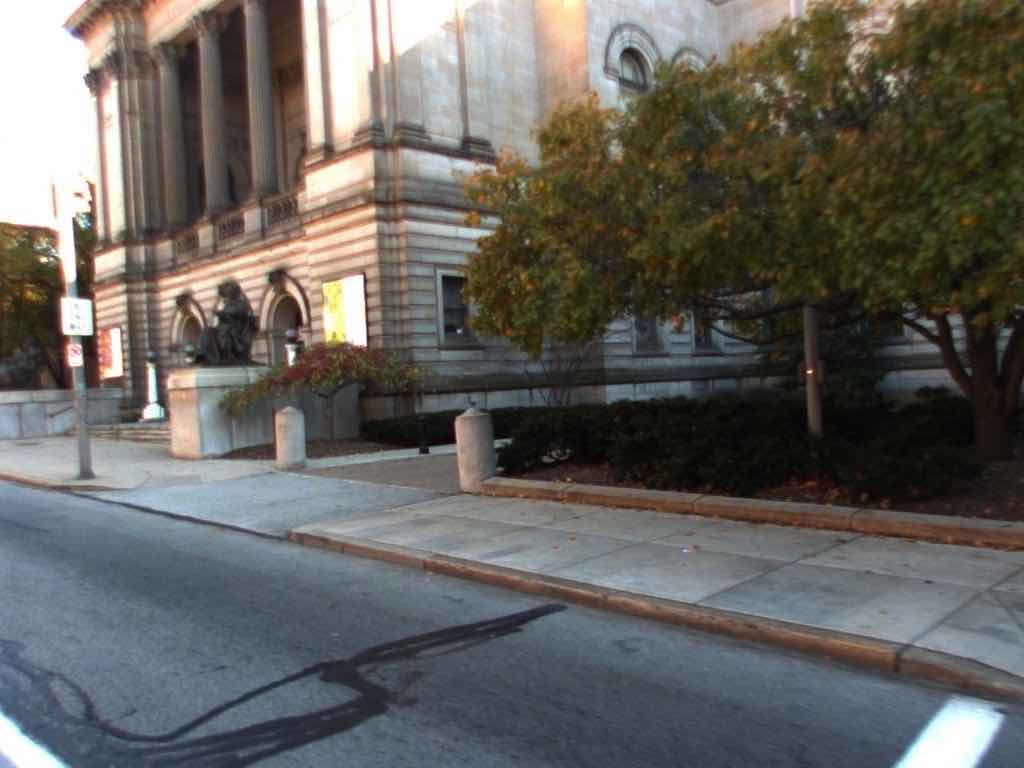}\hfill
	\includegraphics[width=0.16\textwidth]{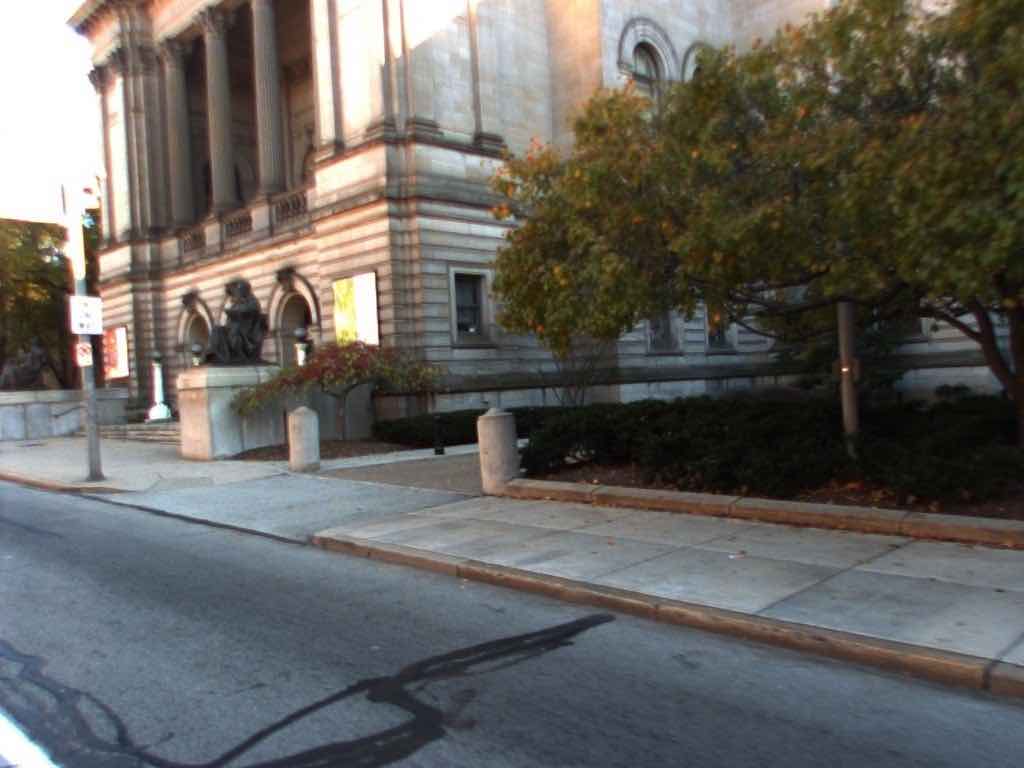}\hfill
	\caption{A subsequence from ``Street Change,'' a proposed dataset for change detection and captioning for visual streams. The dotted line denotes the location of a changepoint, where at least one non-distractor element of the scene has changed. Some human captions for the change across the changepoint are: ``A road sign is gone.'' ``The orange construction sign was removed.'' ``The safety sign is not in the road.''}\label{fig:vlcmu-0}
\end{figure}

In general, visual streams (illustrated in Figure~\ref{fig:vlcmu-0}) consist of time series of images of a scene, such as time lapse imagery or a sequence of satellite images. Given such visual streams, our goal is to detect and describe ``relevant'' changes while ignoring nuisance changes such as lighting or perspective changes or seasonal change. This change detection and description task is relevant to intelligence, insurance, urban planning, and natural disaster relief. 

Our contributions to visual stream change detection and captioning
are:
\begin{enumerate}
	\item We formulate the joint change detection and description
	problem and develop two new datasets, one synthetic and one
	real, of visual streams with known changepoints and change
	descriptions in natural language. %\gscom{rephrased}
	\item We propose a semi-supervised framework for describing
	changes between images that leverages unlabeled image pairs
	to improve performance, an important feature given
	the cost of human labeling of visual streams.
	\item We propose and analyze an approach to change
	detection within a visual stream. It is based on a regularized 
	graph cut cost that also incorporates the
	consistency of perceived change for image pairs straddling the change point. The
	representation used to assess such consistency emerges from
	learning to describe change with language. We show that this approach is
	superior to baselines that do not incorporate language in learning.
\end{enumerate}

The efficacy of including our language model in the change
detection process shows that access to language change \emph{descriptions} at training time may provide a richer form of supervision
than training using the times of changes alone. This is a new, previously unexplored avenue for leveraging the interaction between vision and language.

\section{Related work}\label{sec:background}

While automatic image captioning has received growing attention in the
computer vision community in the past several years
\cite{xu2015show,johnson2016densecap,fang2015captions,ling2017teaching},
such tools are inadequate for  detection of changes in a 
visual stream. Describing {\em changes} in a scene is fundamentally
different from recognizing objects in an image. 
Unlike in classical automatic image captioning, a visual stream may contain no relevant changes and it may contain multiple objects that do not change. 
Furthermore, video captioning methods~\cite{venugopalan2015sequence,pan2016hierarchical,wang2017video} are inapplicable because in many visual streams there are relatively large periods of time between subsequent images; for instance, a satellite may pass over a location only once per day. Finally, the images of
interest may be corrupted by blur, shifts, and varying environmental
conditions that make identifying changes challenging. 

\subsection{Automatic description of the visual world}
\noindent\textbf{Visual descriptions}
The task of describing the content of an image or video in natural
language, often referred to as \emph{image/video captioning}, has emerged as
one of the central tasks in computer vision in recent
years~\cite{vinyals2015show,donahue2015long,johnson2016densecap,venugopalan2015sequence,shetty2016frame,zhou2018end,krishna2017dense}.
Typically,
captioning is approached as a translation task,
mapping an input image to a sequence of words.
In most existing work, learning is fully supervised,
using existing data sets~\cite{young2014image,lin2014microsoft,krishna2017visual,xu2016msr} of ``natural images/videos'' with associated
descriptions. However, when faced with drastically different
visual content, \eg remote sensing or surveillance videos, and a different type of descriptions,
in our case describing change, models trained on natural data lack both the visual
``understanding'' of the input and an appropriate vocabulary for the
output. We address this limitation by
leveraging novel semi-supervised learning methods, requiring only a small amount of labeled
data.

\noindent\textbf{Automatic description of
	change}. Change is implicitly
handled in methods for captioning \emph{video}~\cite{pan2016hierarchical,wang2017video,pan2017video,baraldi2017hierarchical}.
However, when describing change in the videos, these
methods tend to focus on describing the ``updated scene'' rather than the
nature of the change between the scenes.

The most related work to ours is \cite{jhamtani2018learning,park2019robust,oluwasanmi2019fully}, who also 
attempt to generate captions for non-trivial changes in images.
\cite{jhamtani2018learning} proposes the Spot-the-Diff dataset, where pairs of images are 
extracted from surveillance videos. The viewpoints are fixed and there is always some change.
\cite{park2019robust} proposes CLEVR-Change, a dataset where pairs of synthesized images are provided with
their synthesized captions. Lighting and viewpoint can vary between image pairs.
We extend \cite{park2019robust} in two ways. First, we consider a series of images instead of two,
and a change can either happen at any timestep, or not at all. Secondly, we consider a
semi-supervised setting where only a subset of image series has associated captions. Moreover,
we report results not only on CLEVR-change and Spot-the-Diff, but also a newly collected street-view dataset.

\noindent\textbf{Semi/unsupervised captioning}
The goal of unsupervised/unpaired captioning is to learn to caption from a set of images
and a corpus, but with no mapping between the two. Usually, some pivoting is needed:
\cite{gu2018unpaired,gu2019unpaired} tackle unpaired captioning
with a third modality,
chinese captions and scene graph respectively. Similarly, \cite{feng2019unsupervised,laina2019towards}
use visual concept detectors
trained on external data to bridge the unpaired images and captions.

% \cite{chen2017show} adapt their model trained on COCO to unpaired data from a new domain.
In the semi-supervised setting, extra unlabeled data supplement paired image caption data.
\cite{liu2018show} uses a self-retrieval module to mine negative samples from unlabeled
images to improve captioning performance.
\cite{kim2019image} uses a GAN to generate pseudo labels, making extra unpaired images and captions paired.
\cite{anne2016deep,venugopalan2017captioning} exploit additional unpaired text and image for novel object captioning.

Our work is most similiar to \cite{liu2018show} but with several important differences. First, we use a real-fake discriminator instead of a retrieval-based discriminator.
Second, the datasets/tasks are different. Our datasets are more under-annotated and out-of-domain than
COCO\cite{lin2014microsoft}, a large natural image dataset that can benefit
easiliy from pretrained vision networks.
Note that, despite the similar technique, our focus in this paper is not
to propose a new semi-supervised technique, but a more general setting of change descriptions in an annotation-restrained situation.

\subsection{Change point detection and localization}
Change point detection and localization is a classical problem in time
series analysis. Typical approaches aim to determine whether and at what time(s)
the underlying generative model associated with the signal being
recorded has changed~\cite{padilla2019optimal,garreau2018consistent,matteson2014nonparametric}.
Such techniques are also used in video indexing, to detect a scene cut, a shot boundary, etc.
\cite{priya2010video,sze2003scene,irani1998video,smoliar1994content,hu2011survey,yu2001efficient}.

In this work, we focus on finding a point that a nontrivial change in content happens while 
many \emph{nuisance changes}
such as misalignments or shifts between images and illumination
changes, seasonal variations, etc., may dominate the visual changes.
Much current work on high-dimensional change point detection (\eg \cite{cho2015multiple,wang2018high}) assumes that the time series has a mean that is
piecewise-constant over time; this is clearly violated by real world
data, such as shown in Fig~\ref{fig:vlcmu-2}.
We show that learning to describe changes can help locate changepoints.

\section{Semisupervised learning of change description}\label{sec:annotation}
Suppose we have $L$ image pairs \emph{labeled} with
change descriptions, $\left(\ibe^\ell,\iaf^\ell,w^\ell\right),
\ell=1,\ldots,L$. Of these, some may correspond to no change, with
$w^\ell=$ like ``no change'' or ``nothing changed''.
An additional set of $U$ \emph{unlabeled} pairs
$\left(\ibe^u,\iaf^u\right), u=1,\ldots,U$ is presumed to only
contain pairs where a change does occur between $\ibe^u$ and
$\iaf^u$, although no description of this change is available.

\begin{figure}
	\includegraphics[width=\textwidth]{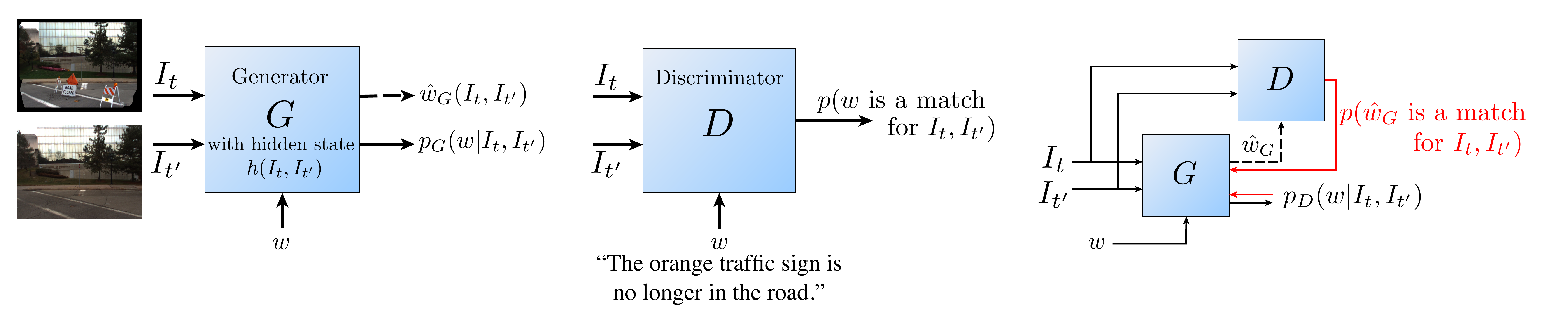}\hfill
	\caption{Description generator and discriminator. $G$ can compute the likelihood for a
		given change caption $w$ for a given pair of images $(I_t,I_{t'})$, as well as
		generate (sample; dashed arrow) a caption $\widehat{w}_G$ for that
		pair. $D$ can estimate the probability that a given
		caption $w$ is a valid description of change between $I_t$
		and $I_{t'}$. During the final phase of training, the generator receives feedback, illustrated in \textcolor{red}{red}, from two sources: ground truth captions, which should have high likelihood under $G$, and the discriminator $D$, which outputs a probability that the sampled captions from $G$ are valid for the pair.
	}
	\label{fig:generator} \label{fig:discriminator}\label{fig:p3}
\end{figure}

\noindent Learning to generate descriptions proceeds in three phases:
\begin{itemize}
	\item {\em Phase 1 (Sec.~\ref{sec:phase1}):} We  use labeled data to
	pretrain a description generator $G$, Fig.~\ref{fig:generator} (left). A conditional generative
	model, $G$ allows us to calculate the likelihood $p(w|I_t, I_{t'})$ of a change description $w$
	given an image pair, as well as sample (estimate) a caption $\hat{w}_G$ for a given
	image pair. The former capacity is used only during training. 
	As part of its operation, $G(\ibe,\iaf)$ computes a
	hidden state $h(\ibe,\iaf)$. 
	\item {\em Phase 2 (Sec.~\ref{sec:phase2}):} We use both labeled and
	unlabeled data to train a discriminator $D$, Fig.~\ref{fig:discriminator} (center).
	For two images $I_t$ and $I_{t'}$ and caption $w$, this
	discriminator computes a ``compatibility score''  $D(w,I_t,I_{t'})$ indicating whether
	$w$ is a valid description of the change between $I_t$ and $I_{t'}$.
	\item {\em Phase 3 (Sec.~\ref{sec:phase3}):} We fine-tune the
	description generator $G$ using both labeled and unlabeled data, leveraging the discriminator $D$, Fig.~\ref{fig:p3} (right). 
\end{itemize}
\noindent Before describing these training phases in detail, we outline the
architecture choices for $G$ and $D$. $G$ uses the DUDA architecture introduced in \cite{park2019robust}, which can be trained
to be robust to lighting and pose changes. Briefly, it extracts visual
features from $I_t$ and $I_{t'}$ using a ResNet~\cite{he2016deep}, and
distills them via spatial attention maps $A_t$ and $A_{t'}$ into a
triplet of visual feature vectors, associated with $I_t$, $I_{t'}$ and the
difference between the two. Finally, these and the caption
$w=(w_1,\ldots,w_K)$ are fed into an LSTM-based
recurrent network with temporal attention states $a_1,\ldots,a_K$ used
to combine the visual features at each time step.

The discriminator $D$ shares the components of DUDA with $G$. However,
instead of using the LSTM model to estimate proability of the word
sequence $w=(w_1,\ldots,w_K)$ given $(I_t,I_{t'})$, here the LSTM model just consumes
the words in $w$, updating the hidden state. The final state $h_K$ is
fed to a multi-layer fully connected network, and finally to a linear binary
classifier. Its output $D(w,I_t,I_{t'})$ is an estimate of the probability that the $w$ is
a valid description of change from $I_t$ to $I_{t'}$.

\subsection{Phase 1: Training description generator $G$}
\label{sec:phase1}

As in~\cite{park2019robust}, training of the description generator $G$ is done with the regularized cross-entropy loss,
\begin{align}\label{eq:generatorlossphase1}
L_G :=  -\sum_{\ell=1}^L \log p(w^\ell | I_t^\ell, I_{t'}^\ell) + \lambda \|A_t^\ell\|_1 + \lambda \|A_{t'}^\ell \|_1 - \mu \sum_{k=1}^{K_\ell} H(a_k^\ell)
\end{align}
where $w^\ell$ is the ground truth change description for
the labeled pair $(\ibe^l,\iaf^l)$, $A_t^\ell$ and $A_{t'}^\ell$ are
the spatial maps computed inside DUDA for that pair, and
$\{a_k^\ell\}$ is the sequence of temporal attention states computed
inside DUDA; $H$ is the entropy. For further details, refer to \cite{park2019robust}.

\subsection{Phase 2: Learning to recognize valid change captions}
\label{sec:phase2}

We train a discriminator using both labeled and unlabeled
image pairs to minimize the loss
\begin{align}
L_D :=& -\sum_{\ell=1}^L \log D(w^\ell, I_t^\ell, I_{t'}^\ell)&
\text{[positives from labeled pairs]}\label{eq:dloss-lpos}\\
& - \sum_{\ell \neq \ell'} \log (1-D(w^\ell, I_t^{\ell'},
I_{t'}^{\ell'}))&\text{[negatives from labeled pairs]}\label{eq:dloss-lneg}
\\
& - \sum_{\ell, u} \log(1-D(w^\ell,I_t^u,I_{t'}^u))
&\text{[negatives across labeled/unlabeled]}\label{eq:dloss-uneg}
\end{align}
The sum in~\eqref{eq:dloss-lpos} is over labeled pairs with known
change annotations; these are the positive examples for $D$. Note that
these may includes pairs with no change, annotated by captions like
$w=$``no change''. The second sum in~\eqref{eq:dloss-lneg} is over
examples obtained by matching one pair with the annotation
for another pair. Here we assume that with high probabilty such a
randomly matched annotation will not in fact be valid for the given
pair, and so these are negative examples for $D$.

Finally, we turn to the unlabeled pairs. The third sum
in~\eqref{eq:dloss-uneg}
matches annotations from labeled data to unlabeled
pairs, which we assume will not create valid
descriptions. This procedure creates additional negative examples for $D$.

We include negative examples from labeled images for two reasons. The first
is to ensure the network is trained using negative examples even in the fully-supervised case.
Secondly, we wish to prevent overfitting, where $D$ learns simply to output 
0 for all pairs in the unlabeled set, and 1 for all of the labeled set.

\subsection{Phase 3: Updating  generator $G$ using discriminator $D$}
\label{sec:phase3}   
With $D$ in hand, we can refine the generator $G$ using both labeled
and unlabeled data. We  train $G$ to generate change captions that the discriminator
labels as ``valid,'' while still staying close to the ground truth data. To ensure that $G$ generates captions that resemble true labels, we simultaneously maximize the likelihood of the ground truth and the outputs of the discriminator for captions generated by $G$. We use the following loss:
\begin{align}
L_3 :=& \sum_{\ell=1}^L L_G(w^\ell,I_t^\ell, I_{t'}^\ell) - \lambda \sum_{\ell=1}^L \log D(G(I_t^\ell,I_{t'}^\ell),I_t^\ell,I_{t'}^\ell) \\
& \qquad - \lambda \sum_{u=1}^U \log D(G(I_t^u,I_{t'}^u),I_t^u,I_{t'}^u) \nonumber 
\end{align}

Since sampling captions from $G$ to pass to $D$ is not differentiable, we treat training $G$ using $D$ as a reinforcement learning problem, which we solve using REINFORCE \cite{williams1992simple}. This approach has been used elsewhere for learning in sequence generation \cite{luo-discriminability-2018,hendricks2016generating}. 

In this case, the policy is given by $G$, parametrized by the network parameters $\theta_G$, which predicts a distribution $p_G(w|I_t, I_{t'})$ over captions. We wish to maximize some reward, $R(w, I_t, I_{t'})$, which should be high for valid change captions and their associated images. The gradient of the expected reward for $G$ is approximated by:
\begin{align*}
\nabla_G \mathbb{E}_{w} \left[ R(\hat{w}_G,I_t, I_{t'}) \right] \approx R(\hat{w}_G,I_t, I_{t'}) \nabla_{\theta_G} \log p_G(\hat{w}_G|I_t, I_{t'}),
\end{align*}
where $\hat{w}_G$ is the result of sampling from $p_G(w|I_t, I_{t'})$, and $p_G(\hat{w}_G|I_t, I_{t'})$ is defined by the output of $G$. In our setting, $R(\hat{w}_G, I_t, I_{t'})$ is given by $\log D(\hat{w}_G,I_t,I_{t'})$. This permits the gradient of $L_3$ to be approximated at training points, and for unlabeled data to be used while training $G$, to ensure that as much data is incorporated into the training procedure as possible.

\begin{figure}
	\begin{tikzpicture}
	\node at (0,0) (I0) {};
	\foreach \in/\ipth in
	{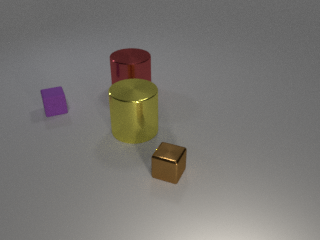, 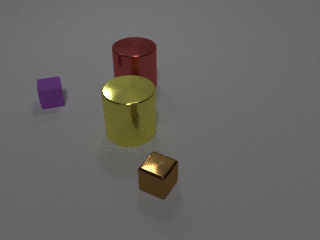, 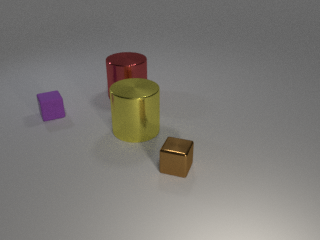,
		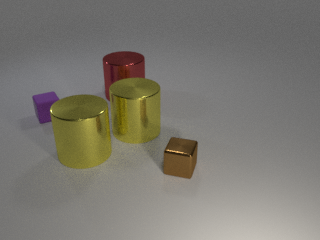, 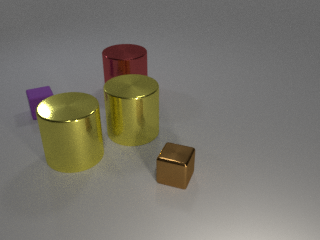}
	{
		\pgfmathsetmacro\iprev{\in -1}
		\node[right=of I\iprev,right=-2pt,label={[black,above=-5pt]above:{$I_\in$}}] (I\in) {\includegraphics[width=.7cm]{\ipth}};
	}
	\draw[ultra thick,black,dashed] ($ (I3.north east)!.5!(I4.north west) $)
	--
	($ (I3.south east)!.5!(I4.south west) $) node[below] (tau)
	{$\tau=3$};
	\node[below=of tau,below=-2pt] {$w=$``A yellow cylinder};
	\node[below=of tau,below=8pt] {appeared by the purple block.''};
	\node[above right=of I5,right=15pt] (p12) {\textbf{($I_1,I_2$,``no
			change'')}};
	\node[below=of p12,below=0pt] (p13) {\textbf{($I_1,I_3$,``no
			change'')}};
	\node[below=of p13,below=0pt] (p23) {\textbf{($I_2,I_3$,``no
			change'')}};
	\node[below=of p23,below=0pt] (p45) {\textbf{($I_4,I_5$,``no
			change'')}};
	
	\node[right=of p12,right=5pt] (p14) {\textbf{($I_1,I_4,w$)}};
	\node[below=of p14,below=0pt] (p15) {\textbf{($I_1,I_5,w$)}};
	\node[below=of p15,below=0pt] (p24) {\textbf{($I_2,I_4,w$)}};
	\node[right=of p14,right=5pt] (p25) {\textbf{($I_2,I_4,w$)}};
	\node[below=of p25,below=0pt] (p34) {\textbf{($I_3,I_4,w$)}};
	\node[below=of p34,below=0pt] (p35) {\textbf{($I_3,I_5,w$)}};
	\end{tikzpicture}
	\caption{Mining of a visual stream from CLEVR-Sequence ({\bf left}) with known change point $\tau$ and change annotation $w$, to create labeled image pairs ({\bf right}).}
	\label{fig:streamtrain}
\end{figure}

While GAN training traditionally uses feedback from a learned discriminator as a training signal for a generator, typically GAN training includes optimization over a minimax objective, where the discriminator and generator compete over the course of training. We do not update the discriminator in an adversarial fashion, instead freezing it during the final phase of training.

\subsection{Training with visual streams}
The above relies on {\em pairs} of images. More generally, we may have
access to a training set of {\em visual streams}, which
consist of ordered image sequences $\{I_{t}\}_{t=0}^T$,
associated times $\tau \in \{1,\ldots,T\}$ at which a change occurs,
and (possibly) natural language descriptions of the changes. Such
sequences may be used to generate training data as follows. Consider
an annotated stream $(I_1,\ldots,I_T, \tau, w)$. For every
$t\le\tau,t'> \tau$ it yields an annotated pair $(I_t,I_{t'},w)$, while for
every $t,t'\le\tau$, we get $(I_t,I_{t'},\text{``no change''})$, and
similarly for every $t,t'>\tau$. See Figure~\ref{fig:streamtrain}.

If the stream $(I_1,\ldots,I_T, \tau)$ is not annotated with change
description, we still can get the pairs annotated with ``no change''
for the frame indices that do not straddle $\tau$. However, for
$t\le\tau,t'>\tau$ we can only get unlabeled pairs $(I_t,I_{t'})$.

\section{Change Detection}\label{sec:detection}

The system described in Section~\ref{sec:annotation} can be used not only to {\em caption} 
changes in visual streams, but also to {\em detect and localize the occurence of changes}.
In this section, we describe how to use the learned modules from Section \ref{sec:annotation}, which operate in a pairwise manner, to identify changepoints in image sequences. 

More formally, assume we are given a visual stream, represented as an ordered sequence of images $\{I_{t}\}_{t=0}^T$. Nuisance changes, such as viewpoint, lighting, or weather variation, are observed between each sequential pair of frames of the stream. There is some unknown $\tau \in \{1,\ldots,T\}$ such that between $t=\tau-1$ and $t'=\tau$ a relevant (non-nuisance) change occurs, after which only nuisance changes occur between frames. Our goal in change detection is to estimate $\tau$. 

We begin by defining an abstract pairwise relation $P(I_t, I_{t'})$: $P(I_t, I_{t'})$ is a relation that, given two images $(I_t, I_{t'})$, outputs a statistic which is high for pairs $(I_t, I_{t'})$ where a change has occurred between $t$ and $t'$, and is low otherwise.

We can now use these pairwise statistics to define different {\em score functions} for various candidate values of $\tau$ in a visual stream; in general, we will compute an estimate $\hat \tau$ by maximizing these score functions. For any method which generates a score $S_X$, $\tau_X = \argmax_{\tau} S_X$.

\begin{figure}
	\begin{minipage}[c]{.5\linewidth}
		\includegraphics[width=\linewidth]{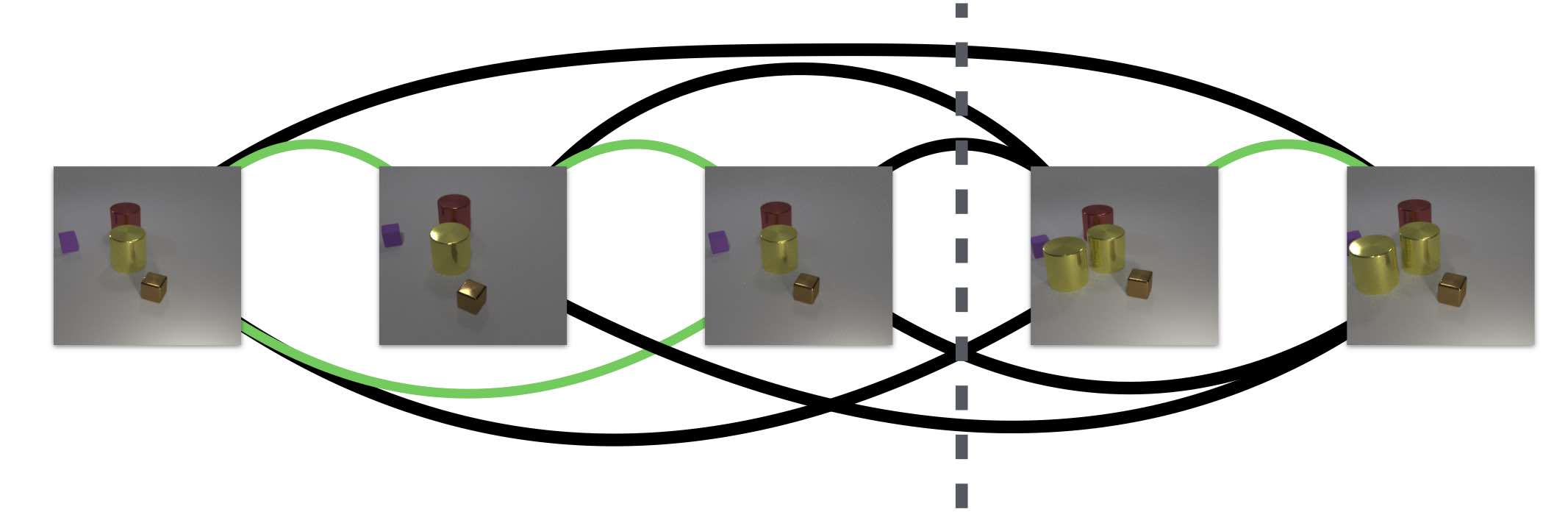}   
	\end{minipage}%
	\begin{minipage}[c]{.5\linewidth}
		\caption{Visual representation of 
			graph given by a time series. Selecting a changepoint $\tau$
			(denoted by the dotted line), partitions the edges into
			$E_\tau$, in black, and \textcolor{Green}{$E^c_\tau$, in
				green}.}
		\label{fig:changepointfiggraph}
		\label{fig:changepointfig}          
	\end{minipage}
	
\end{figure}

\paragraph{Step-wise scores.}
We define the step-wise score as
\begin{equation}\label{eq:stepwisechangepoint}
S_{\rm step}(\tau) := P(I_{\tau-1},I_\tau) 
\end{equation}
and the associated change point estimate as $\tau_{\rm step} = \argmax_{\tau} S_{\rm step}(\tau)$.
This approach finds the consecutive pair of frames in the stream which is most likely to contain a change between that pair. Unfortunately, if $P(I_{t}, I_{t'})$ is noisy or imperfect, the step-wise  approach may be prone to errors and spurious detections. 

\paragraph{Graph cut scores.} To address the noise sensitivity for the step-wise approach, we propose a strategy based on graph cuts. Specifically, we form a graph with each node corresponding to a $t \in \{1,\ldots,T\}$. We assign an edge weight to each pair of nodes $t$ and $t'$ (specifics below) and define a score function with the edge weights.
Selecting a candidate change point $\tau$ is then equivalent to selecting a graph cut, where the edges that are cut connect some $t \le \tau$ and $t' > \tau$; 
we denote the set of these edges as $E_{\tau}$. A visualization is provided in Figure \ref{fig:changepointfiggraph}, where a candidate $\tau$ results in an edge partitioning into $E_\tau$, in black, and \textcolor{OliveGreen}{$E^c_\tau$, in green}.

The graph-cut score we consider is, for $\textcolor{OliveGreen}{E^c_{\tau}}$ the complement of $E_{\tau}$,
\begin{equation}\label{eq:graphcutcost}
S_{\textrm{GC}} = \frac{1}{|E_{\tau}|} \sum_{(t, t') \in E_{\tau} } P(I_t, I_{t'})  - \textcolor{OliveGreen}{ \frac{1}{|{E_{\tau}^c}|} \sum_{(t, t') \in E^c_\tau } P(I_t, I_{t'})}.
\end{equation}

\paragraph{Computing Change Statistics.} In order to compute the pairwise statistic $P(I_t, I_{t'})$, we explore ways to use the networks $D$ and $G$ introduced in Section~\ref{sec:annotation}, as well as a standalone, ``image-only'' network trained specially for the task.
First, we note that by fixing the test caption to the ``no change'' token, we can define:
\begin{equation}\label{eq:pairwisediscstatistic}
P_D(I_t, I_{t'}) := - D(\textrm{``no change''}, I_t, I_{t'})
\end{equation}
which uses the discriminator's label as a measure of how poorly the ``no change'' description fits the pair of images $I_t$ and $I_{t'}$.

We also define $\tilde D(I_t, I_{t'})$, an ``image-only" change detector. $\tilde D(I_t, I_{t'})$ learns to produce visual features using the same encoder architecture $E_{t,t'}$ as the generator in Section~\ref{sec:phase1}. The visual features are passed through several fully-connected layers, and the output of $\tilde D(I_t, I_{t'})$ is a single label, rather than a caption. $\tilde D(I_t, I_{t'})$ is trained to classify whether pairs contain a change or not using a binary classification loss, where a label of 1 indicates no change. 
Now we can define:
\begin{equation}
\tilde P(I_t, I_{t'}) := -\tilde D(I_t, I_{t'}),
\end{equation}
so $\tilde P(I_t, I_{t'})$ is an estimate of the probability that a change has occurred between $t$ and $t'$ that has been trained {\em without using language information}. We define an ``image-only'' stepwise score $S_{\rm step-IO}$ where $\tilde P$ is used for $P$. Similarly, we also define an ``image-only'' variant of the graph cut score of $S_{\rm GC-IO}$ using $\tilde P$.

\paragraph{Consistency-regularized graph cut scores.}

The above step-wise and graph cut change detection scores depend only on the pairwise relation $P(I_t, I_{t'})$. However, we may have access to learned hidden representations $h(I_t, I_{t'})$ that contain more information relevant to change detection, which may help regularize the above graph cut scores. Specifically, we regularize our score function to help ensure that all the pairs $(I_t,I_{t'})$ with $t \le \tau$ and $t' > \tau$ have similar representations (since they should represent the same change). We measure the similarity between representations of two pairs, $h(I_t,I_{t'})$ and $h(I_{s},I_{s'})$ using the cosine similarity.

This leads to a regularized ``representation consistency'' score  of 
\begin{align}
\label{eq:RC}
S_{\textrm{RC}} =& \argmax_{\tau}  \frac{\lambda}{|E_{\tau}|} \sum_{(t, t') \in E_{\tau} } P(I_t, I_{t'}) -\textcolor{OliveGreen}{\frac{\lambda}{|{E^c_{\tau}}|} \sum_{(t, t') \in E^c_{\tau} } P(I_t, I_{t'})} \\ 
&\qquad  + \sum_{(t, t') \in E_{\tau}, (s, s') \in E_{\tau}} \cos\left[h(I_t, I_{t'}), h(I_s, I_{s'}) \right] \nonumber
\end{align}
%\becca{here $h$ is a function of $I_t$ and $I_{t'}$, but in section 3 it's $h(w,I_t,I_{t'})$ -- which is right?}
for a user-specified tuning parameter $\lambda \ge 0$; then $\hat \tau_{\rm RC} := \argmax_\tau S_{\rm RC}$. The consistency measure in the final term of \eqref{eq:RC} sums over all \emph{pairs} of black edges in Figure~\ref{fig:changepointfiggraph}, finding the average similarity of the associated hidden representations.

To make the above concrete, we may choose $P$ to be $P_D$ from Equation~\ref{eq:pairwisediscstatistic}, and let the representation of a pair $(I_t,I_{t'})$ be the final hidden state of the caption generator $G(I_t, I_{t'})$. In this formulation, a natural interpretation of the estimator arises: we would like to determine a changepoint $\hat{\tau}$ where for all pairs $(I_t,I_{t'})$, where $t \le \tau$ but $t' > \tau$, $D(\text{``no change''}, I_t, I_{t'})$ is low. In addition, the representation consistency pushes us to choose a changepoint where the generator $G$ will output similar descriptions of the change that has occurred across $\tau$ for all possible pairs. In this formulation, the language model becomes an integral part of the changepoint detection method.

Absent a language model, we can also define an ``image-only representation consistency''. To do this, in Equation~\ref{eq:RC} we utilize $\tilde P$, and for internal representation $\tilde h(I_t, I_{t'})$ we use the penultimate layer of the image pair used by $\tilde D(I_t, I_{t'})$, a configuration that computes $S_{\rm RC-IO}$.

\section{Data}\label{sec:data}

The experiments in Section~\ref{sec:experiments} consider both (a) an assessment of our proposed system of change captioning and the role of unlabeled data in the performance, and (b) an evaluation of our approach when both annotated and unannotated visual streams are available for change annotation and detection. The datasets used for these experiments are described below.

\subsection{Semi-supervised Change Captioning}\label{sec:semisup}

CLEVR-Change is a simulated dataset, with a variety of distractor changes and a curated set of ``relevant'' changes, first presented in \cite{park2019robust}. Each image consists of colored geometric primitives on a flat, infinite ground plane. The images are generated with the CLEVR engine \cite{johnson2017clevr}, a standard tool in a variety of visual reasoning tasks. \clevrchange~includes, for each initial frame, a frame in which only nuisance changes have occurred, as well as a frame in which both nuisance and relevant
changes have occurred. 
There were initially roughly 39000 change/distractor pairs in the CLEVR-change dataset, which we have augmented with 10000 additional unlabeled pairs of images by mimicking their generation process. The additional unlabeled pairs always contain a change.

Spot-the-Diff \cite{jhamtani2018learning} is a video surveillance change description dataset which consists of pairs of surveillance images and associated human-generated captions for the visual changes in each pair. Each pair is aligned and assumed to contain a change, with similar lighting conditions between the ``before'' and ``after'' frames, so this dataset does not contain the nuisance changes present in the \clevrchange~dataset. Spot-the-Diff contains 13192 image pairs, which we split into 10000 training pairs and 3192 testing pairs.

\subsection{Visual Streams}

Due to the lack of standard datasets for our task, we propose two datasets for evaluation of captioning and changepoint detection on visual streams.

The first dataset we introduce is a modification of the \clevrchange~synthetic dataset. By modifying the machinery in \cite{park2019robust}, we generate a visual stream instead of a pair of images. Over the course of each sequence, the camera follows a random walk from its initial location, while the light intensity also follows a random walk, to ensure that the proposed methodology must be robust to nuisance changes. At some time in the sequence, one of the non-nuisance changes used in the original \clevrchange~dataset occurs. Each sequence contains 10 images, with the changepoints  uniformly distributed in $t\in\{1,\ldots,8\}$ (assuming zero indexing).  Further details can be found in the supplementary materials. Figure~\ref{fig:streamtrain} shows a subsequence drawn from one CLEVR-Sequence stream.

We also propose the ``Street Change'' dataset, which consists of 151 image sequences captured in an urban environment during different seasons and times of day, extracted from the VL-CMU dataset \cite{badino2012real}. Each sequence contains multiple views of a particular scene that are captured at an initial date, and then a second subsequence that is captured at a later date. The original change detection dataset was curated by 
\cite{alcantarilla2018street}, and we have utilized human labelers on Amazon's Mechanical Turk to provide change annotations for training. The annotation set includes at least three captions for both the original and time-reversed sequences, which were curated after data collection to ensure high-quality labels.

The ``Street Change'' dataset contains both minor and major nuisance changes across the dataset. First, from an annotation perspective, each frame is captured at a different viewpoint, and there are often lighting changes across the sequence. Moreover, since some sequences span seasons, environmental changes like plant growth, snowfall, and weather variation are present in the dataset, which any successful trained model must learn to ignore. An example sequence is shown in
\ref{fig:vlcmu-2}, along with the associated human annotations. More examples are included in the supplementary materials. The average length of each stream is 9 images, the mean sentence length is 6.2 words with standard deviation 1.9, and the maximum sentence length is 16 words. There are 420 unique words in the captions.

\section{Experiments}\label{sec:experiments}

In this section we empirically demonstrate the performance of the proposed method on both caption generation and changepoint detection. 
We find that the proposed training regime improves on the state of the art for change captioning, as well as an adapted semi-supervised approach from standard image captioning. 

We find that the RC approach, which leverages both $D$ and the generative language model $G$, outperforms our other proposed changepoint detection methods. While adding the representation consistency term also improves ``Image-Only'' approaches, we show that incorporating both $D$ and $G$ results in good performance on language metrics for change captioning, as well as superior change detection performance..
\subsection{Implementation Details} $G$ is trained as in Phase 1 for 40 epochs using Adam \cite{adam}, and the discriminator $D$ is also trained for 40 epochs using Adam. Phase 3 of training proceeds for 20 epochs. The learning rate begins at 0.001 and decays by a factor of 0.9 each 5 epochs. $\lambda$ in $L_3$ is set to be 0.2.
The image-only discriminator $\tilde{D}$ is trained for 40 epochs using Adam with a learning rate of 0.001.

$G$ is identical in architecture to the network used in \cite{park2019robust} for all three training sets. The three architectures differ only in the dimension of the final output of $G$, $p_G(w|I_t, I_{t'})$, which is 64, 2404, and 420 for CLEVR, Spot-the-diff, and Street Change respectively. At test time, sampling captions is done in a greedy fashion. 
\paragraph{Adapting Show, Tell, and Discriminate to change captioning.} An alternative approach to recognizing valid descriptions is proposed in \cite{liu2018show} for single-image captioning. Their approach, called Show, Tell, and Discriminate (ST\&D),  embeds the descriptions $w$ and images $I$ in a common space, and then measures the similarity of the embeddings by measuring the cosine similarity $s(w, I)$ between them. The embedding module is trained with a hard triplet loss on labeled data, of the form: $L_{\textrm{ST\&D}} = \sum_{i=1}^L \max_{i\neq j} \left(\alpha - s(w^i, I^i) + s(w^i, I^j) \right)_{+}$
for margin parameter $\alpha$. 
Training in stage three proceeds similarly to our approach, with $L_{\textrm{ST\&D}}$ providing a form of discriminative reward to $G$.

We have modified the \std approach to perform change captioning, so the embedding module operates on pairs of images rather than single images. In Section \ref{sec:experiments} we compare the performance of the modified \std to our approach. The learned joint embedding space is dimension 2048, and the margin used for the triplet loss is 0.1. We follow the authors' lead and set the weighting of the self-retrieval module in stage three to be 1.

\begin{table}
	\centering
	\resizebox{\columnwidth}{!}{
	\begin{tabular}{c|c|c|c|c|c|c|c|c|c|c|c|c|c}
		\hline
		\multicolumn{2}{c|}{Labeled} & \multicolumn{3}{c|}{1000} & \multicolumn{3}{c|}{2000} & \multicolumn{3}{c|}{10000} & \multicolumn{3}{c}{20000}  \\ \hline
		\multicolumn{2}{c|}{Unlabeled} & 1k & 10k & 30k  & 1k & 10k & 30k  & 1k & 10k & 30k  & 1k & 10k & 30k \\ \hline \hline
		\multirow{4}{*}{Ours} & C & 57.1 & 44.8 & 42.7 & 58.1 & 60.2 & 61.4 & 93.7 & 98.5 & 103.3 & 100.2 & 100.5 & 110.0   \\  
		& B & 30.8 & 28.5 & 26.9 & 34.2 & 34.6 & 34.7 & 49.3 & 51.5 & 51.6 & 54.5 & 54.8 & 56.4  \\  
		& M & 32.2 & 30.1 & 29.5  & 32.4 & 32.6 & 33.2 & 39.3  & 39.3  & 39.5 & 39.9 & 41.1 & 42.9  \\  
		& R & 69.2 & 65.6 & 65.1 & 70.2 & 70.3 & 70.9 & 78.8  & 79.8 & 80.8 & 81.2 & 82.4 & 83.5  \\  \hline
		\multirow{4}{*}{ST\&D} & C & 42.4 & 42.7 & 43.6 & 55.7 & 58.2 & 60.3 & 92.8 & 93.9 & 102.0 & 98.4 & 105.1 & 106.4   \\  
		& B & 27.0 & 26.3 & 27.1 & 32.4 &  46.9 & 47.2 & 48.7 & 51.8 & 52.0 & 55.3 &  55.3 & 57.3  \\  
		& M & 29.4 & 29.8 & 29.6  & 31.5 & 32.8 & 33.0 & 38.3  & 38.7  & 40.7 & 38.9 & 40.9 & 42.2  \\  
		& R & 65.4 & 65.8 & 65.6 & 65.8 & 69.8 & 70.0 & 78.9  & 79.1 &  81.5 & 79.3 & 82.0 & 83.3 \\  \hline
	\end{tabular}
	}
	\caption{Semi-supervised scaling with respect to labeled and unlabeled training set size for CLEVR-Change using our method and Show, Tell, and Discriminate. We report CIDEr (C), BLEU-4 (B), METEOR (M), and ROUGE-L (R) (scaled by $100\times$).}
	\label{table:clevrlanguagemetrics}
\end{table}

\begin{table}
	%	\begin{minipage}[c]{.68\linewidth}
	\resizebox{\columnwidth}{!}{
		\begin{tabular}{c|c|c|c|c|c|c|c|c|c|c|c||c |  c | c|c|cc}
			\hline
			\multicolumn{2}{c|}{Labeled} & \multicolumn{4}{c|}{2000} & \multicolumn{3}{c|}{5000} & \multicolumn{2}{c|}{8000} & 9000 & \multicolumn{2}{c|}{L} & 30 & 100 & 130  \\ \hline
			\multicolumn{2}{c|}{Unlabeled} & 1k & 2k & 5k  & 8k & 1k & 2k  & 5k & 1k & 2k  & 1k & \multicolumn{2}{c|}{U} & 100 & 30 & 0 \\ \hline \hline
			\multirow{4}{*}{Ours} & C & 11.1 & 11.4 & 12.5 & 13.2 & 21.7 & 21.8 & 23.1 & 30.4 & 30.8 & 33.3 & \multirow{4}{*}{Ours} & C & 86.6 & 116.0 & 119.8    \\  
			& B & 5.2 & 14.4 & 13.3 & 14.6 & 10.5 & 11.7 & 15.2 & 9.9 & 12.8 & 16.1 & & B & 30.5 & 41.7 & 42.6  \\  
			& M & 19.6 & 22.1 & 22.7  & 23.1 & 22.2 & 22.1 & 23.5  & 22.9  & 23.0 & 24.0 & & M & 33.8 & 40.5 & 39.9    \\  
			& R & 39.1 & 47.7 & 48.6 & 50.3 & 46.2 & 47.5 & 50.4  & 48.5 & 49.3 & 51.4 & & R & 66.6 & 75.9 & 73.5  \\  \hline
			\multirow{4}{*}{ST\&D} & C & 10.2 & 9.9 & 10.5 & 11.4 & 18.5 & 19.6 & 21.3 & 29.0 & 30.1 & 30.8 & \multirow{4}{*}{ST\&D} & C & 64.4 & 105.2 & 112.7    \\  
			& B &  3.8 & 8.4 & 10.1 & 9.6 & 11.2 & 12.9 & 13.7 & 10.6 & 11.5 & 13.0 & & B & 26.7 & 38.9 & 38.6   \\  
			& M & 16.9 & 16.6 & 17.1  & 17.1 & 20.4 & 21.2 & 22.3  & 22.5  & 22.9 & 23.5 & & M & 32.0 & 39.3 & 38.8 \\  
			& R & 38.9 & 38.7 & 40.6 & 41.1 & 42.4 & 43.1 & 47.7  & 47.0 &  47.1 & 47.7 & & R & 64.5 & 72.6 & 73.1   \\  \hline
		\end{tabular}
	}
	%	\end{minipage}%
	\caption{{\bf Left}: Semi-supervised scaling with respect to labeled and unlabeled training set size for Spot-the-Diff, and {\bf Right} Street Change on a variety of language metrics (scaled by $100\times$) using our method and Show, Tell, and Discriminate. Note that training set size for Street Change is in terms of number of labeled sequences, instead of image pairs.}
	\label{table:streetandspotthedifflanguage}
\end{table}

\paragraph{Evaluation} For language tasks, we report BLEU-4 \cite{papineni2002bleu}, CIDEr \cite{vedantam2015cider}, ROUGE-L \cite{lin2004rouge}, and METEOR \cite{denkowski2014meteor}. To evaluate the quality of change detection methods, we generate precision-recall curves; we report full precision-recall curves in Figure~\ref{fig:changepointfiggraph}, as well as summarizing with Average Precision in Table \ref{table:changedetectionboth}.

\paragraph{Semisupervised Change Captioning} We begin by comparing our semisupervised approach to Show, Tell, and Discriminate (ST\&D), a method for semisupervised single-image captioning, which does not utilize unlabeled data to train the network that provides feedback to $G$. Table \ref{table:clevrlanguagemetrics} explores a range of training set sizes, varying both $L$ and $U$, the number of labeled and image pairs respectively. Except in the case of extremely scarce labeled data ($L=1$k), we see consistent improvements from adding unlabeled data; our approach appears to use additional unlabeled data more effectively, overtaking the adapted \std approach.

\begin{figure*}
	\hfill
	\begin{subfigure}[t]{0.49\textwidth}
		\includegraphics[width=\linewidth]{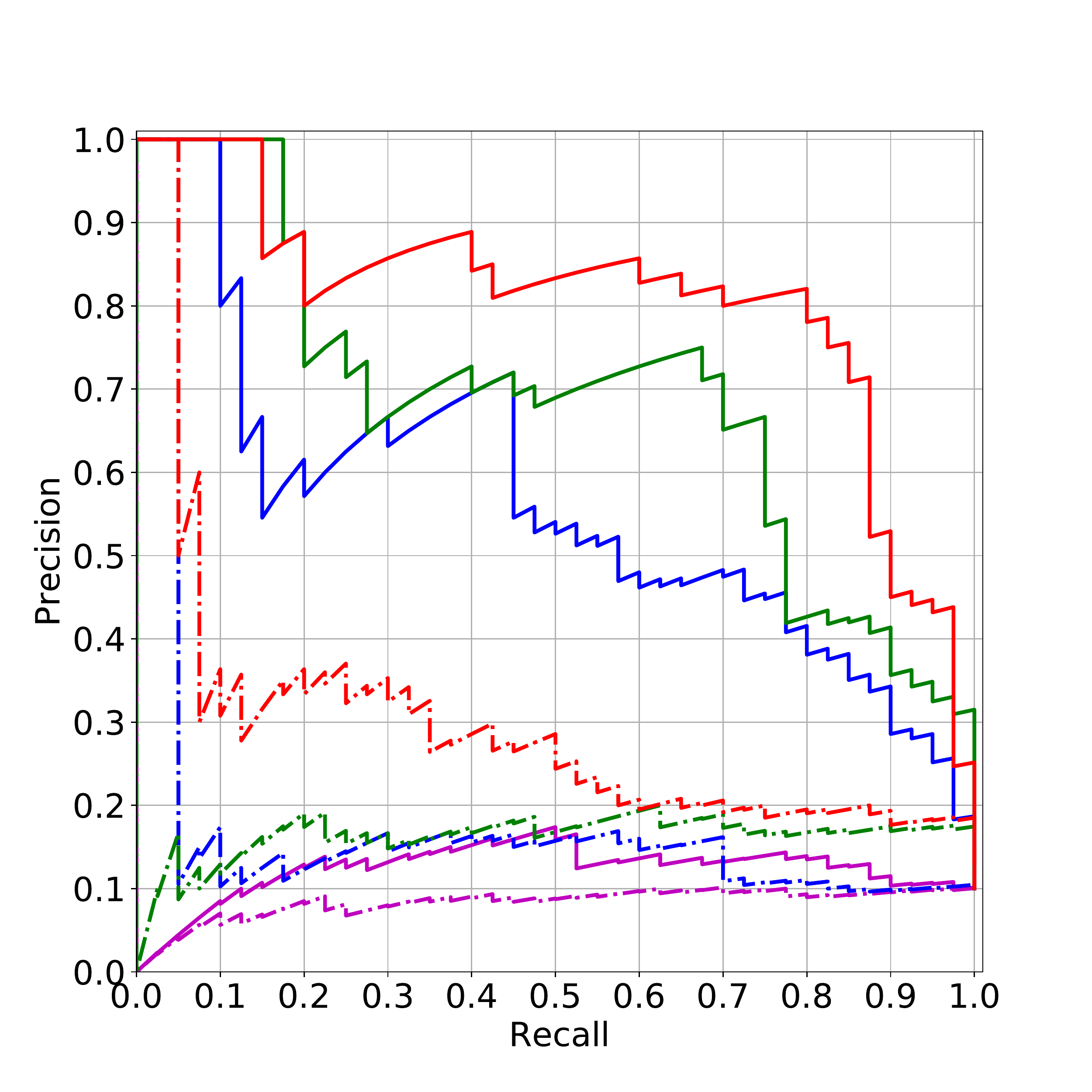}
		%		\caption{}
	\end{subfigure}
	\hfill
	\begin{subfigure}[t]{0.49\textwidth}
		\includegraphics[width=\linewidth]{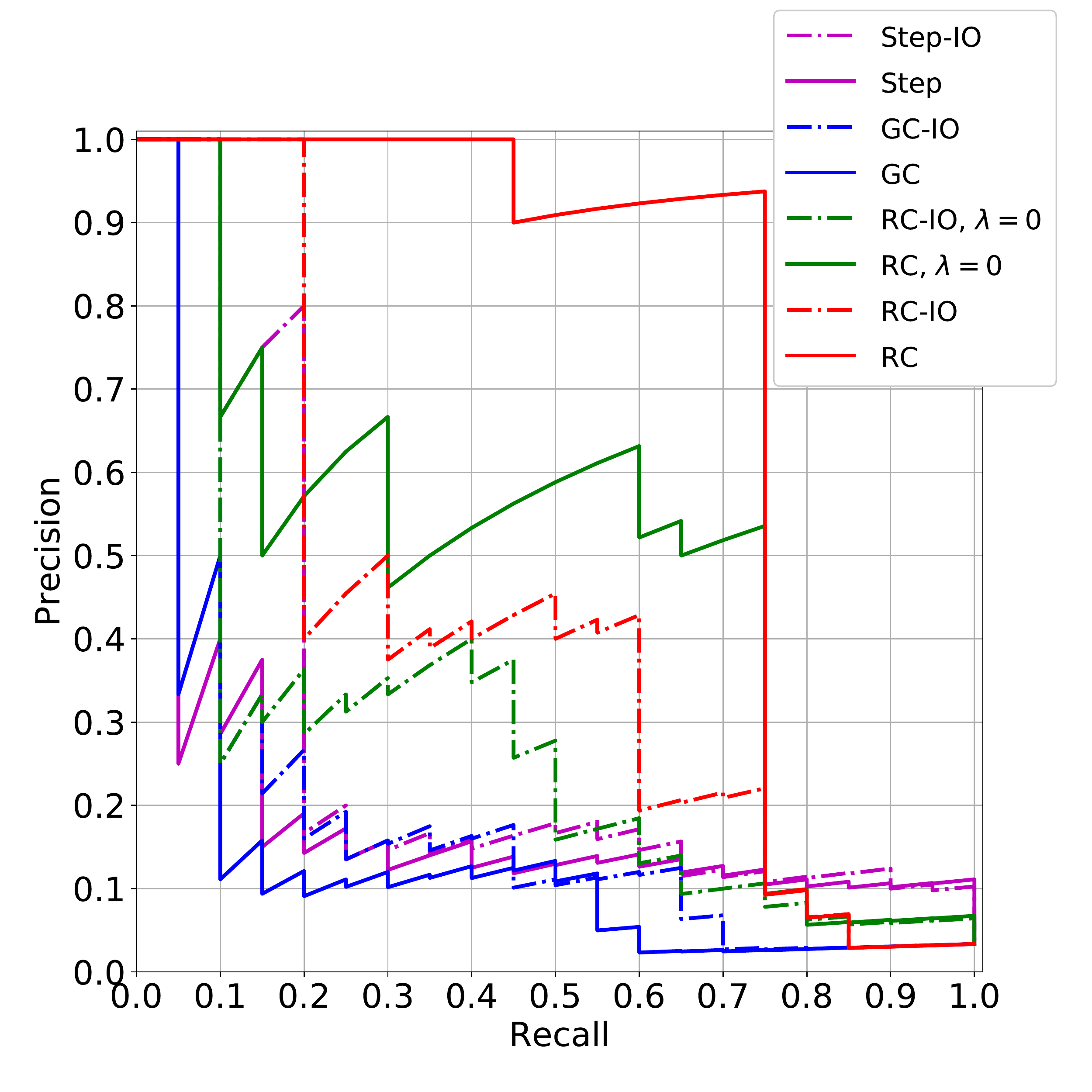}
		%		\caption{}
	\end{subfigure}
	\hfill
	\caption{Precision-recall curve for (a) CLEVR-Sequence, and (b) Street Change. Methods compared are \textcolor{Plum}{Stepwise with Language (Step)}, \textcolor{Plum}{Stepwise with Images Only (Step-IO)}, \textcolor{Emerald!70!black}{Regularized Cut with $\lambda=0$}, \textcolor{Emerald!70!black}{Regularized Cut with Images Only with $\lambda=0$ (RC-IO $\lambda=0$)}, \textcolor{blue}{Graph Cut (GC)}, \textcolor{blue}{Graph Cut with Images Only (GC-IO)}, \textcolor{red}{Regularized Cut (RC)}, and \textcolor{red}{Regularized Cut with Images Only (RC-IO)}. 
	}
	\label{fig:precisionrecall}
\end{figure*}

Table \ref{table:streetandspotthedifflanguage} compares captioning performance with respect to different labeled and unlabeled training set sizes for both Spot-the-Diff (left) and Street Change (right). Note the training set sizes for Street Change are in terms of \emph{sequences}. 
\paragraph{Change Detection.} In Figures \ref{fig:precisionrecall} a) and b), we illustrate the performance of all changepoint methods introduced on both CLEVR Sequence and Street Change. For all methods with the ``Image Only'' prefix ``-IO,'' we utilize $\tilde{P}$ to calculate scores, and for hidden representations $\tilde h(I_t, I_{t'})$ we utilize the penultimate layer of $\tilde D(I_t, I_{t'})$. For methods without the ``-IO'' prefix, we use $P_D$, which utilizes the learned change caption discriminator, and the hidden state of the caption generator $G$ as $h(I_t, I_{t'})$.

\begin{figure}
	\begin{subfigure}{0.24\linewidth}
		\centering
		\includegraphics[width=\linewidth]{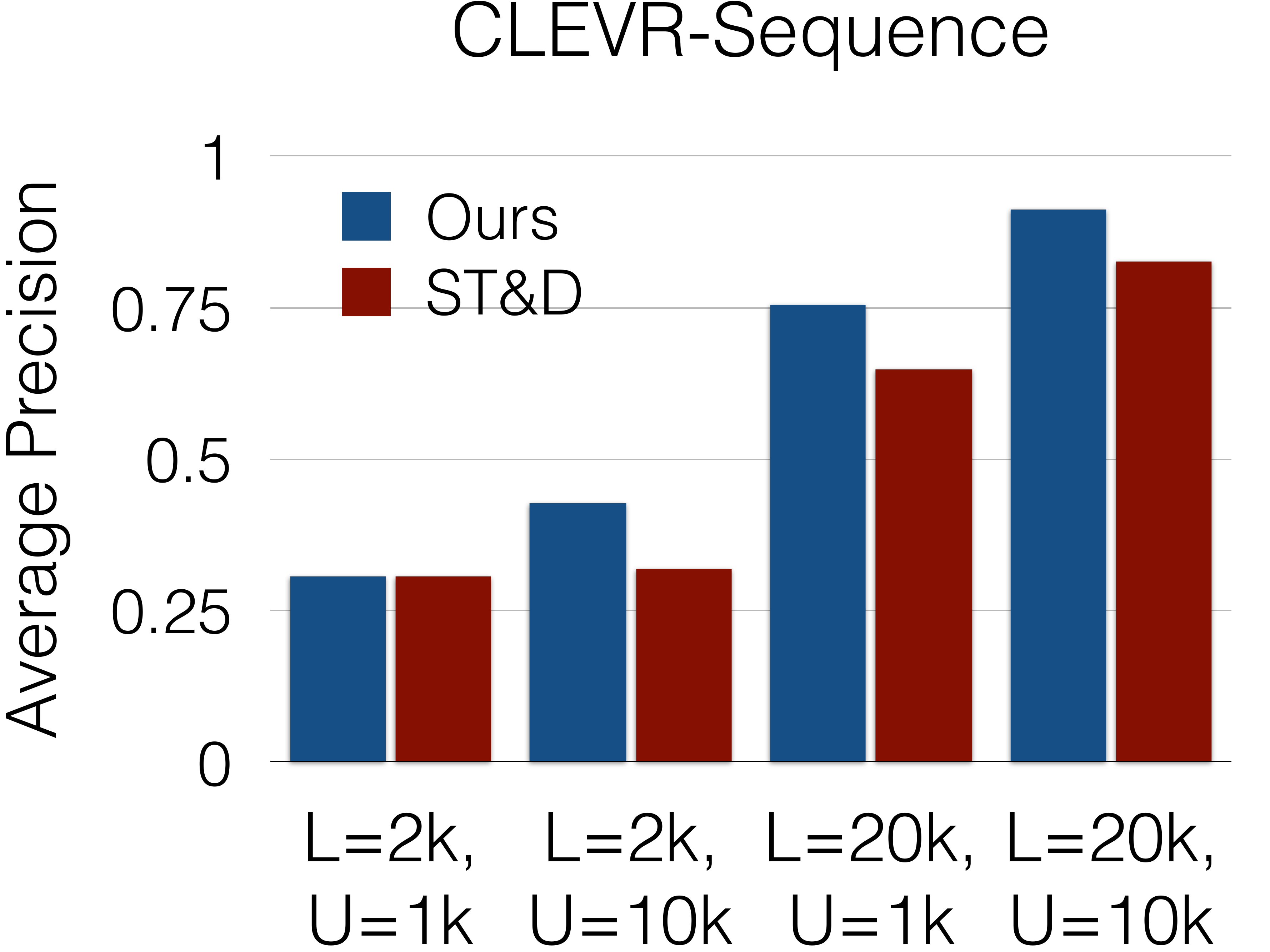}
		\caption{}
		\label{fig:apclevr}
	\end{subfigure}
	%	\hfill
	\begin{subfigure}{0.24\linewidth}
		\centering
		\includegraphics[width=\linewidth]{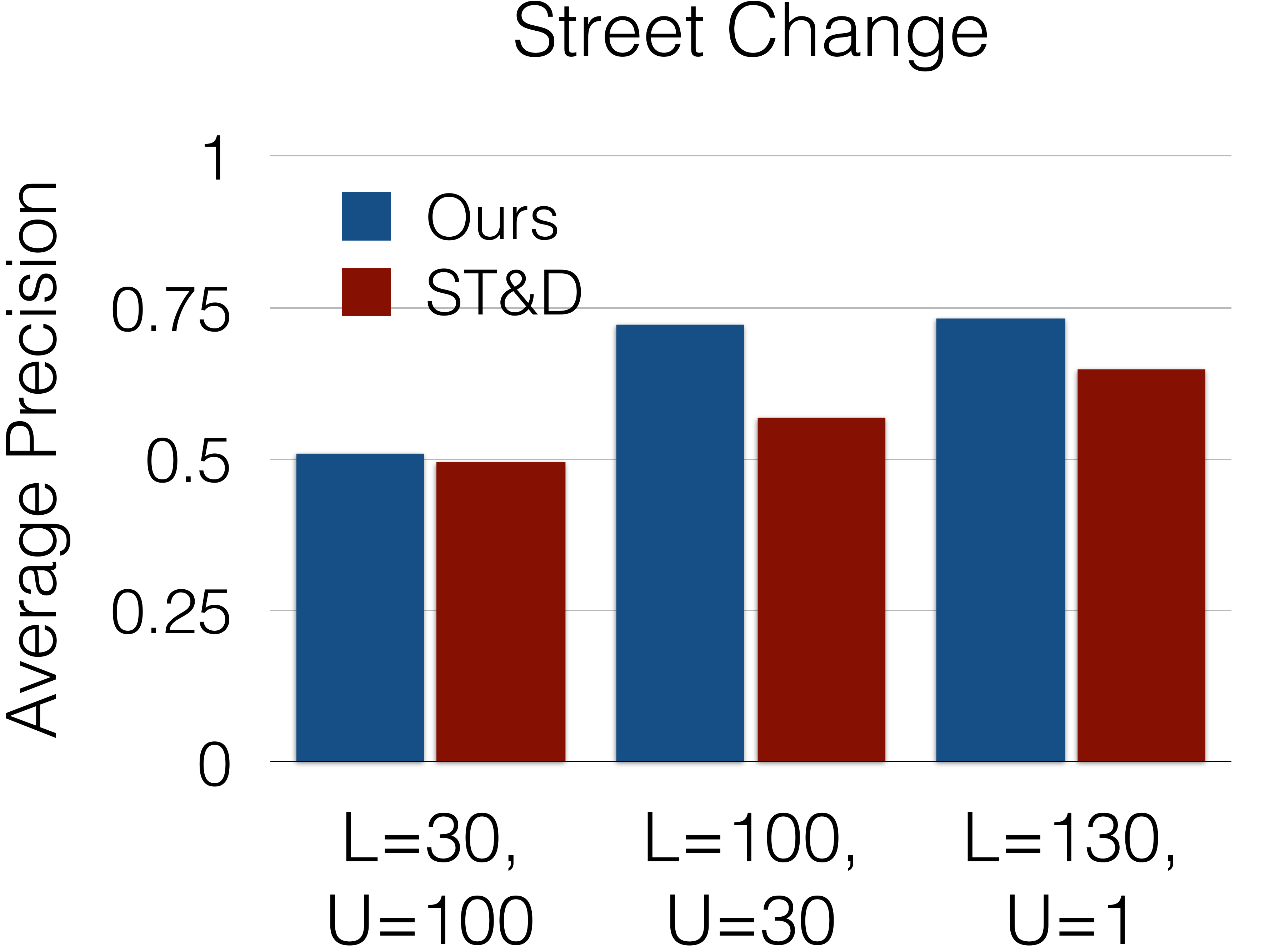}
		\caption{}
		\label{fig:apstreet}
	\end{subfigure}
	\begin{subfigure}{0.24\linewidth}
		\centering
		\includegraphics[width=\linewidth]{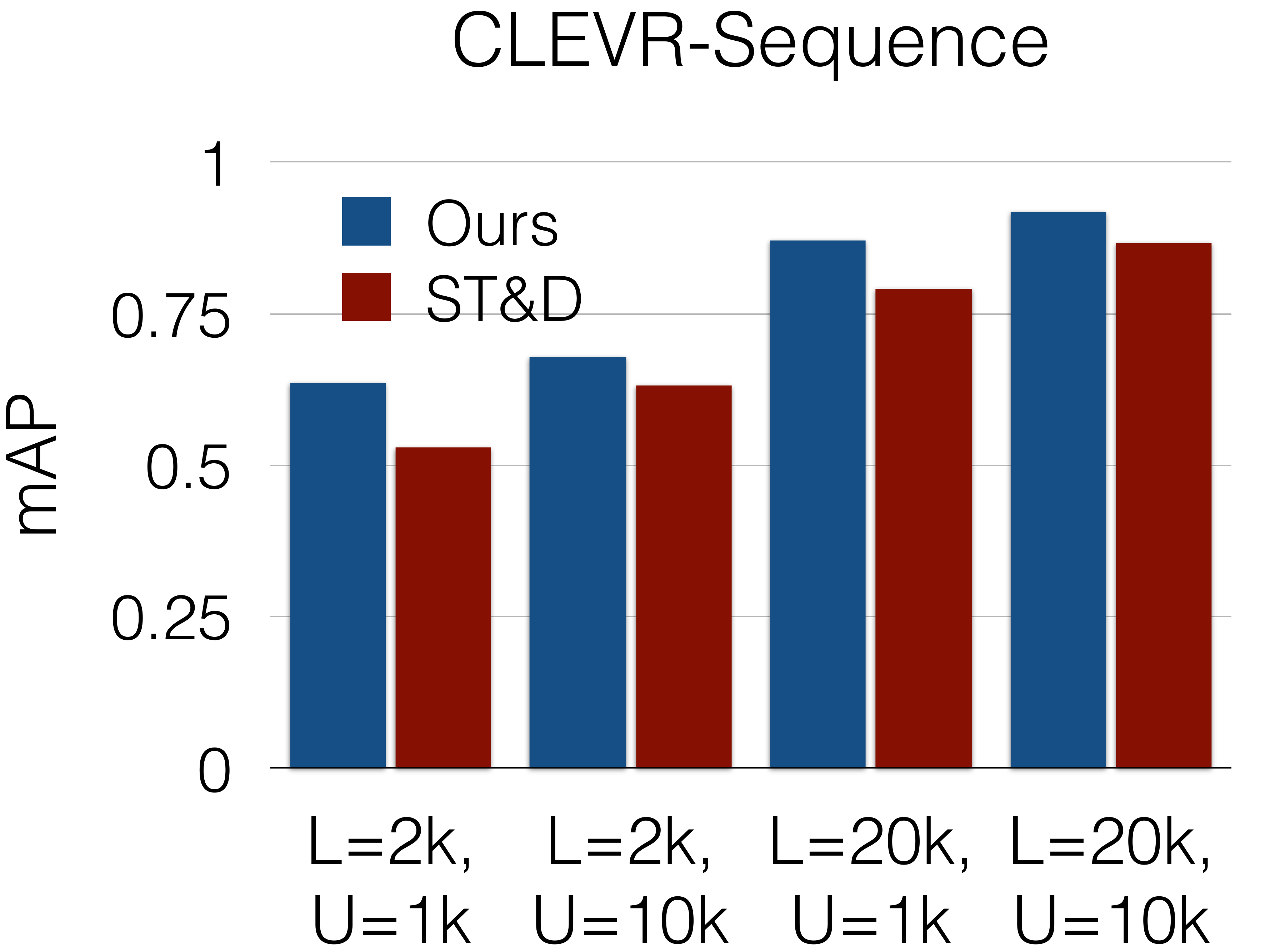}
		\caption{}
		\label{fig:mapclevr}
	\end{subfigure}
	\hfill
	\begin{subfigure}{0.24\linewidth}
		\centering
		\includegraphics[width=\linewidth]{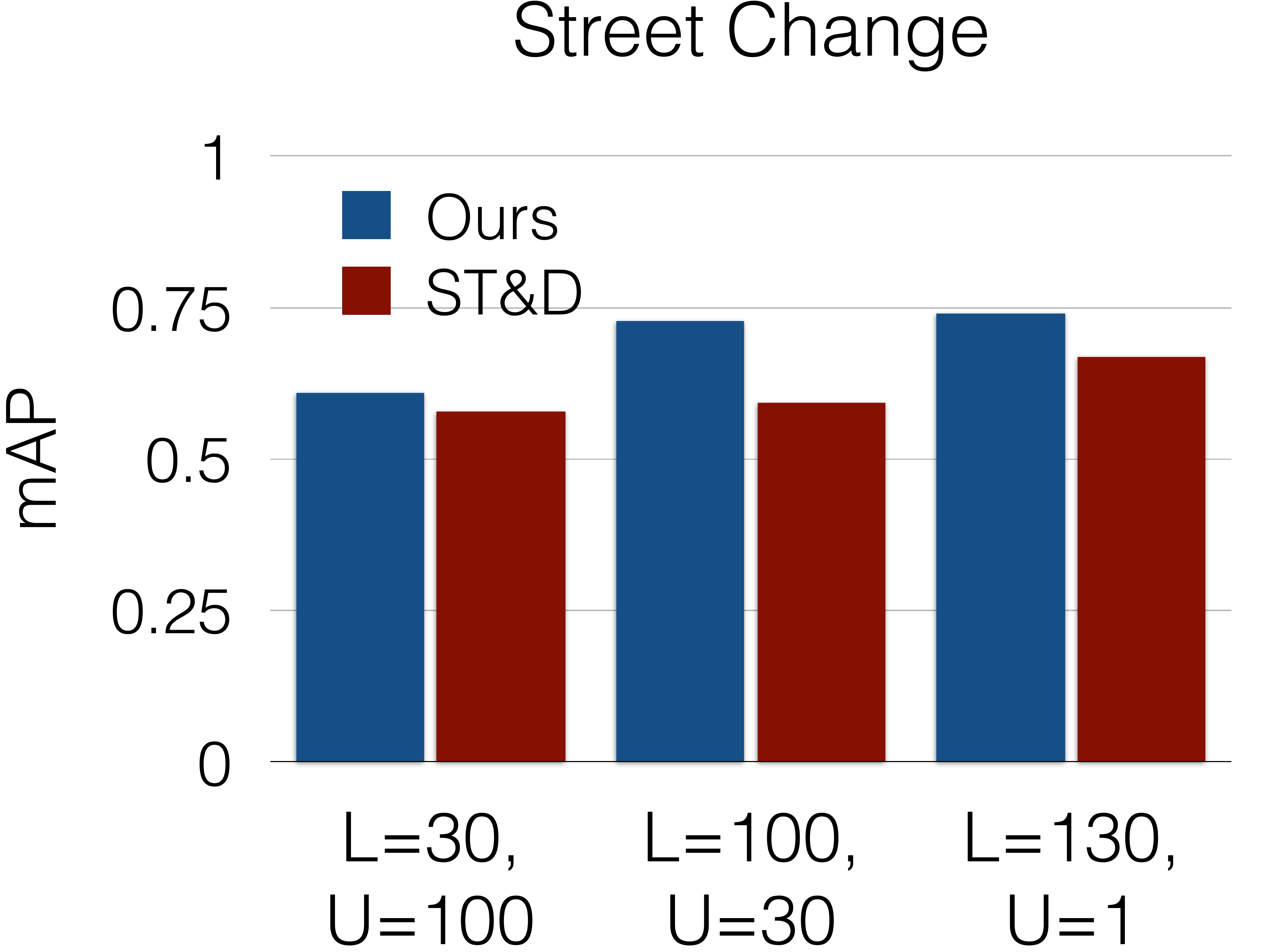}
		\caption{}
		\label{fig:mapstreet}
	\end{subfigure}
	\caption{Relationship between amount of labeled and unlabeled data in the training set and change detection performance for CLEVR-Sequence  and Street Change.}
	\label{fig:samplecompdetection}
\end{figure}

To choose $\lambda$ in $S_{\rm RC}$, a held-out set of 20 sequences from CLEVR Sequence were used. We use $\lambda=5/4$ in both CLEVR Sequence and Street Change tests. We also explore setting $\lambda=0$ to isolate the performance of the representation consistency loss as a standalone method.

Figure \ref{fig:samplecompdetection} illustrates the effect of adding both labeled and unlabeled data to the training set of $G$ and $D$. 
%Adding unlabeled data to the training set improves change detection performance as well as caption quality. 
We observe a consistent improvement from image-only methods to their counterparts that use a language-informed $D$. Graph-cut and representation consistency terms both improve detection accuracy compared to stepwise approaches; combining the two leads to further improvements.

In Table \ref{table:changedetectionboth}, we explore  the effect of loosening the requirements for a ``correct'' change detection. In these cases, we vary a window size parameter $\Delta$, so that a detection is called correct if the absolute error $|\hat{\tau} - \tau^*| \leq \Delta$ is smaller than $\Delta$. $\Delta=0$ is the default for all other tables and figures. 

\begin{table}
	\centering
	\resizebox{\columnwidth}{!}{
	\begin{tabular}{c|c|c|c|c|c|c||c|c|c|c|c|c}
		\hline
		& \multicolumn{6}{c||}{CLEVR-Sequence} & \multicolumn{6}{c}{Street Change} \\ \hline
		Window Size & 0 & 1 & 2 & 3 & 4 & mAP & 0 & 1 & 2 & 3 & 4 & mAP \\ \hline \hline 
		Step-IO & 17.2 & 21.8 & 33.0 & 37.6 & 65.8 & 35.1 & 35.7 & 43.1 & 46.3 & 46.3 & 48.4 & 44.0 \\ 
		Step & 20.8 & 24.9 & 37.1 & 40.5 & 62.3 & 37.1 & 23.9 & 35.6 & 37.7 & 38.9 & 39.0 & 35.0 \\ \hline
		GC-IO & 22.0 & 27.3 & 36.4 & 42.4 & 49.2 & 35.5 & 27.1 & 34.0 & 35.0 & 35.1 & 35.5 & 33.3   \\ 
		GC & 58.4 & 63.9 & 69.0 & 71.1 & 73.2  & 67.1 & 19.8 & 27.1 & 28.8 & 29.9 & 30.8 & 27.3 \\ \hline
		RC-IO $\lambda=0$ & 24.8 & 31.8 & 39.4 & 45.9 & 51.3 & 38.6 & 35.3 & 36.1 & 36.2 & 36.2 & 36.2 & 36.0 \\ 
		RC $\lambda=0$ & 68.8 & 72.2 & 73.1 & 74.4 & 76.6  & 73.1 & 52.2 & 52.8 & 52.9 & 52.9 & 52.9 & 52.7 \\ \hline
		RC-IO & 33.1 & 38.5 & 42.9 & 48.9 & 59.5 & 44.6 & 47.1 & 48.3 & 48.3 & 48.3 & 48.4 & 48.1  \\
		RC & \textbf{79.5} & \textbf{84.0} & \textbf{89.8} & \textbf{91.3} & \textbf{92.9} & \textbf{87.5} & \textbf{72.1} & \textbf{73.2} & \textbf{73.2} & \textbf{73.2} & \textbf{73.2} & \textbf{73.0}  \\  
		\hline 
	\end{tabular}
	}
	\caption{Average Precision (AP) for CLEVR-Change with differing window sizes. Window size represents the maximum distance from the true changepoint for a point to be counted as "correct." mAP is the mean AP over all window sizes. To generate Average Precision (AP) values, we average the precision of a method for recall=$[0.0,0.1,0.2,\ldots,1.0]$.}
	\label{table:changedetectionboth}
\end{table}

\section{Conclusions}\label{sec:conclusion}

In this work, we explore natural language captioning and detection of change in visual streams, illustrating that learning to generate captions to describe change also enhances our ability to accurately detect visual changes over streams. While natural language labels have a strong positive impact on change detection performance, we recognize that labeled training data is often difficult and expensive to obtain. With this challenge in mind, we develop a semi-supervised training regimen that fully leverages unlabeled data. A broad array of experiments on newly developed datasets consisting of visual streams with natural language labels help quantify the performance gains associated with different amounts of unlabeled data on both captioning and detection.

\bibliographystyle{splncs04}
\bibliography{change_description_arxiv}
\newpage
\appendix
\appendixpage

In the following sections we provide further information regarding the ``Street Change'' and CLEVR-Sequence datasets. We provide several example sequences, explain the annotation gathering procedure for Street Change, and provide further information that was not contained in the main document.

\section{CLEVR-Sequence}

CLEVR-Sequence is an extension of the CLEVR-Change dataset of \cite{park2019robust}. The CLEVR-Change dataset consists of triplets of images along with annotations: one ``original'' image, a ``change'' image, and a ``distractor'' image. 

CLEVR-Sequence, by contrast, consists of sequences of ten images. The changes between all but one pair in the sequence are ``distractor'' changes, in which the camera moves and the lighting changes slightly. For one pair, there is, in addition to the distractor change, a non-distractor change, in which one of the possible changes from \cite{park2019robust} occurs with uniform probability over the change types. For clarity, these are: Color (an object changes color), Texture (the material of a single object changes), Add (a new object appears), Drop (a previously-present object disappears), and Move (a single object moves in the scene).

We generate 400 sequences for each candidate changepoint in the range $\{1,2,3,4,5,6,7,8 \}$, assuming 0 indexing. We also generate 400 ``distractor'' sequences in which no change occurs at any point.

We attempt to imitate the characteristics of distractor changes as outlined by the authors of \cite{park2019robust} as closely as possible. Annotations are produced automatically for each sequence using the same methodology as the CLEVR-Change dataset.

\section{Street Change}

The images in the Street Change dataset were curated by \cite{alcantarilla2018street}, and are grouped into 151 unique image sequences of variable lengths. The minimum sequence length is 2 (this length-2 sequence was omitted from training and testing), and the maximum is 42. As mentioned in the main body, the average sequence length is approximately 9.

In Street Change, the change point is always the center frame of the sequence. For methods which learn by leveraging the entire sequence simultaneously, this may be a problematic bias, but since we learn and infer using only pairwise comparisons, this bias is less concerning.

Captions were gathered using Amazon's Mechanical Turk. Annotators were presented with a pair of images and asked to describe in a single brief sentence the major changes between them, ignoring nuisance changes. Annotators were given guidance on what to consider nuisance changes, like weather, lighting, time of day, and camera angle. Annotations for both the original sequences and time-reversed sequences were gathered, as a form of data augmentation.

The gathered annotations were curated by hand to remove descriptions of nuisance changes, ensure consistent spelling, ensure each annotation was a single sentence, and confirm accuracy of the annotations. We collected six annotations for each sequence (and six more for the time-reversed sequence), to ensure that there would be at least three acceptable annotations for each sequence. 

\begin{figure}
	\includegraphics[width=0.19\textwidth]{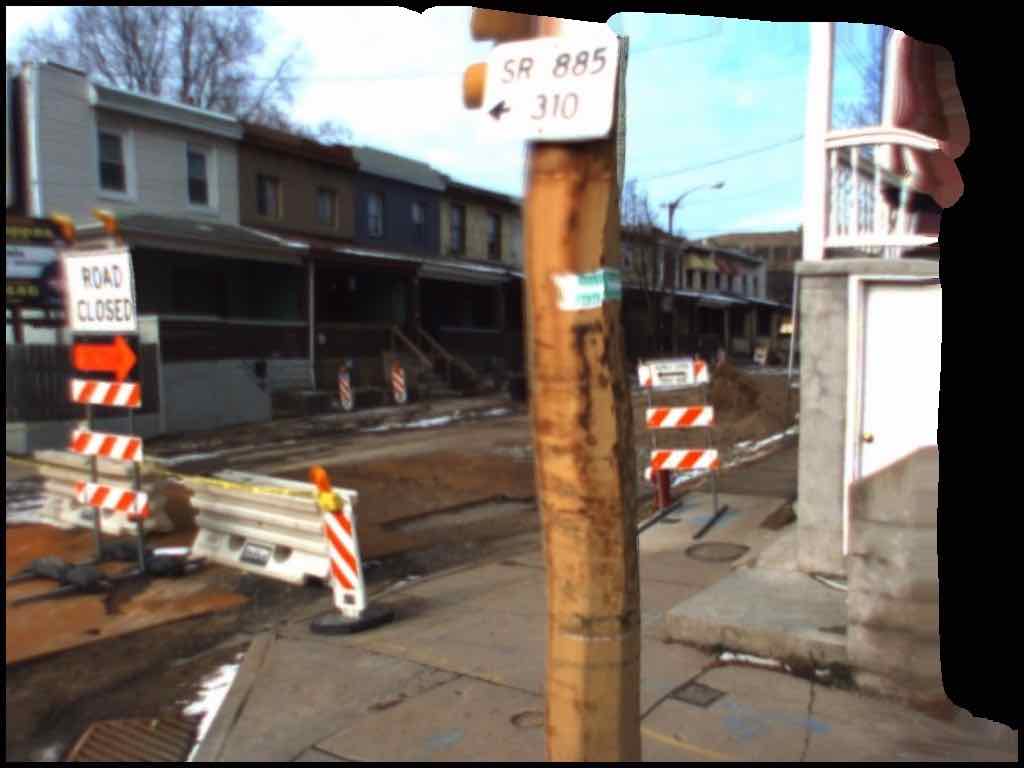}\hfill
	\includegraphics[width=0.19\textwidth]{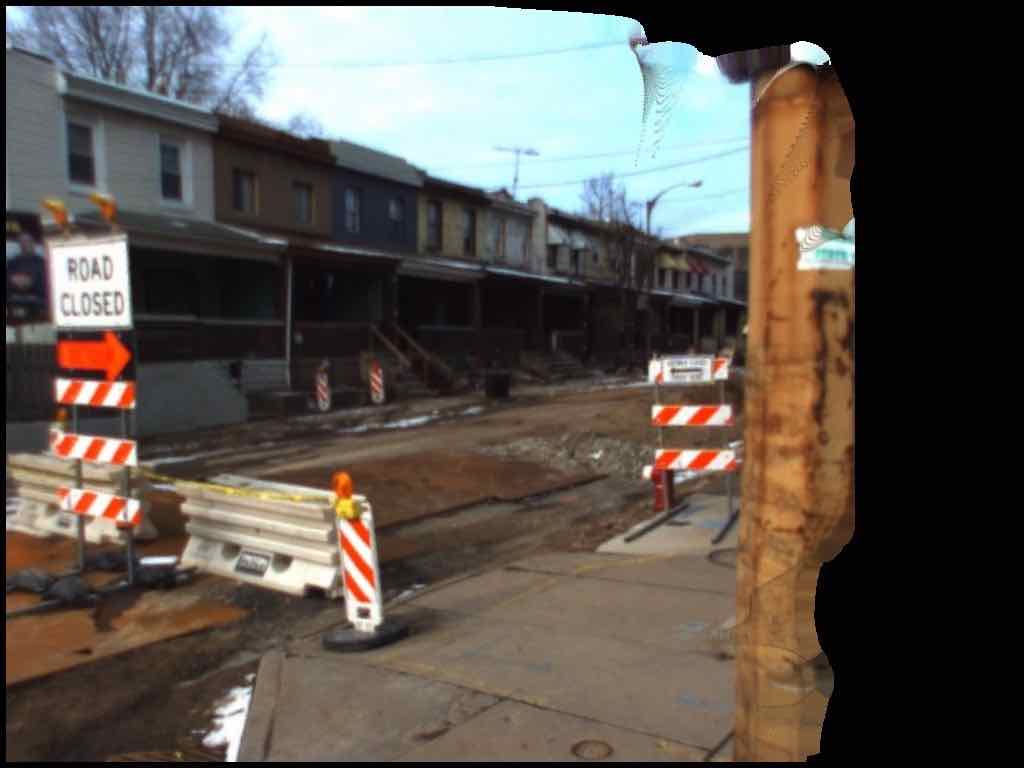}\hfill
	\includegraphics[width=0.19\textwidth]{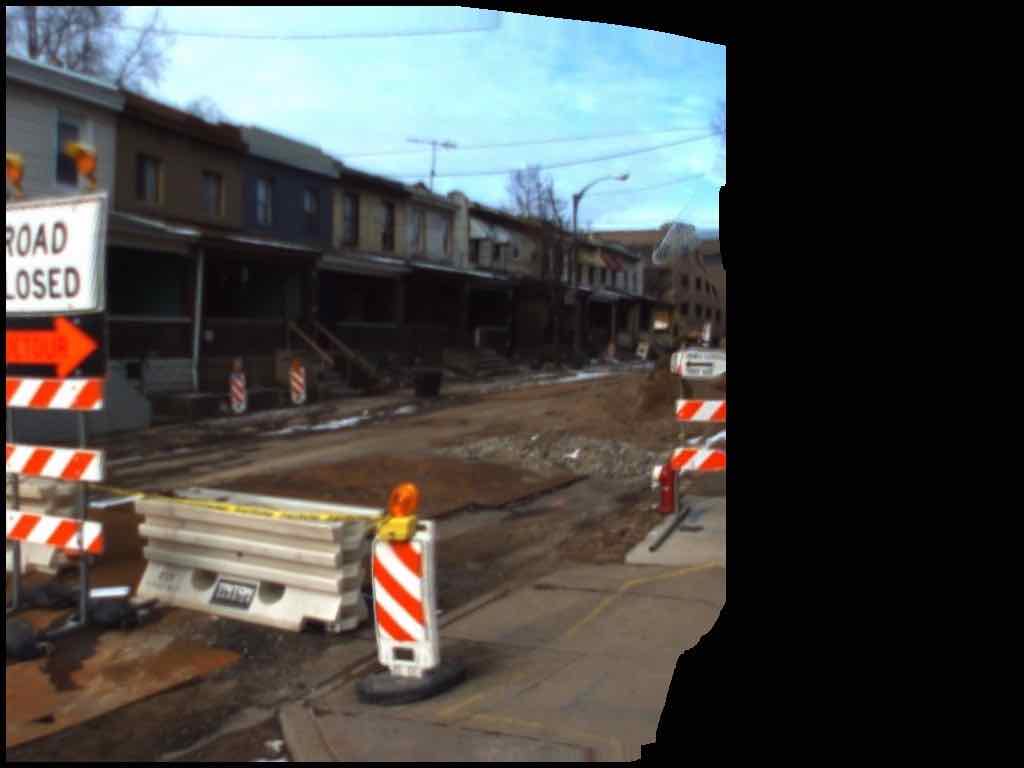}\hfill
	\includegraphics[width=0.19\textwidth]{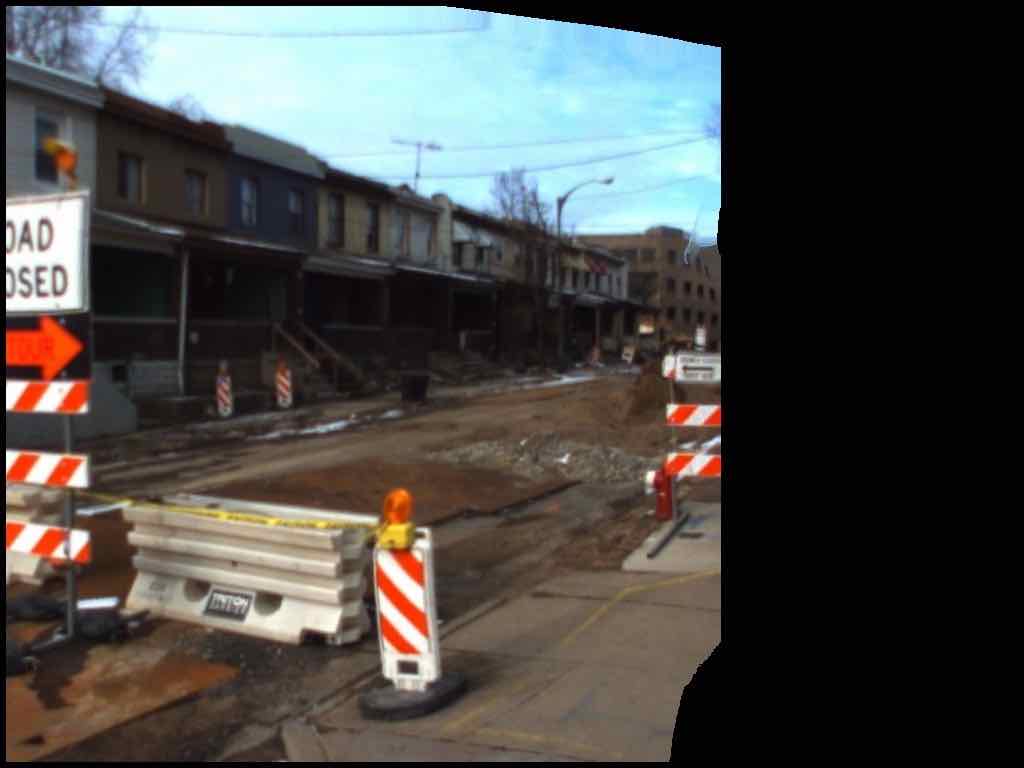}\hfill
	\includegraphics[width=0.19\textwidth]{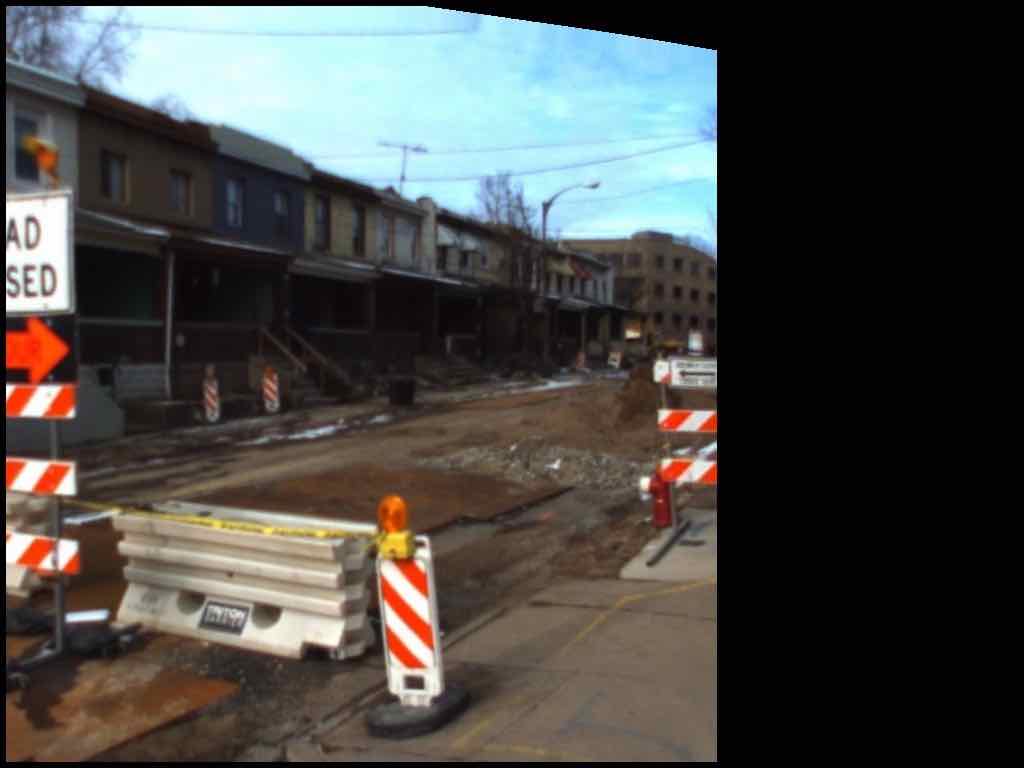}\hfill \\
	\includegraphics[width=0.19\textwidth]{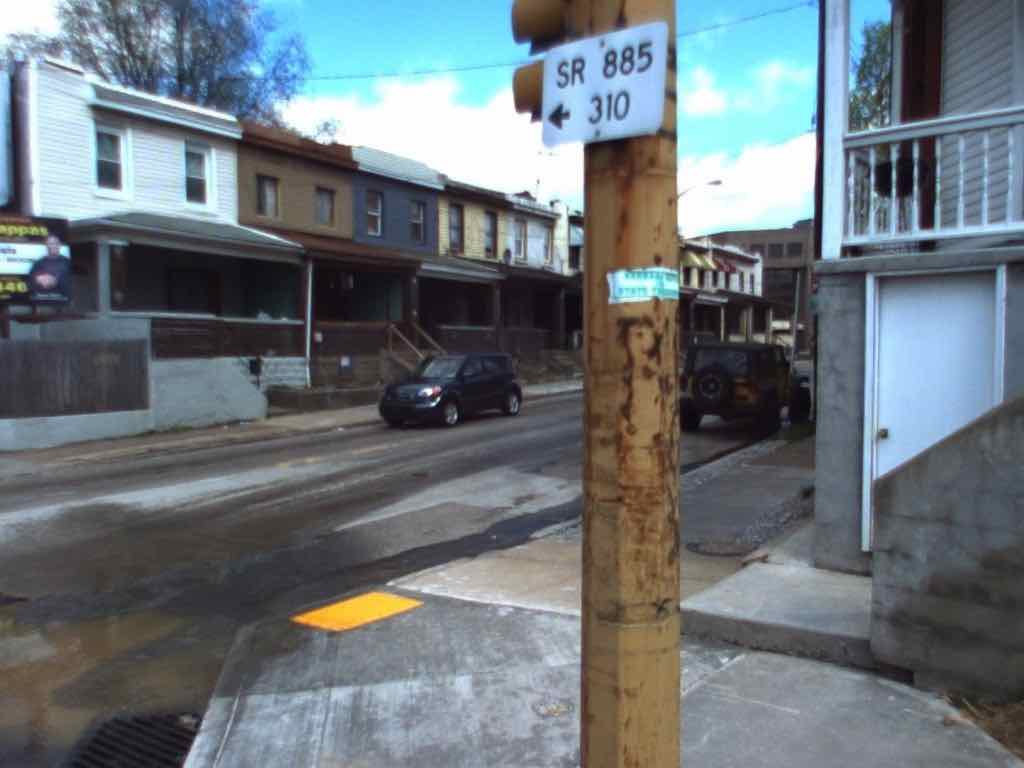}\hfill
	\includegraphics[width=0.19\textwidth]{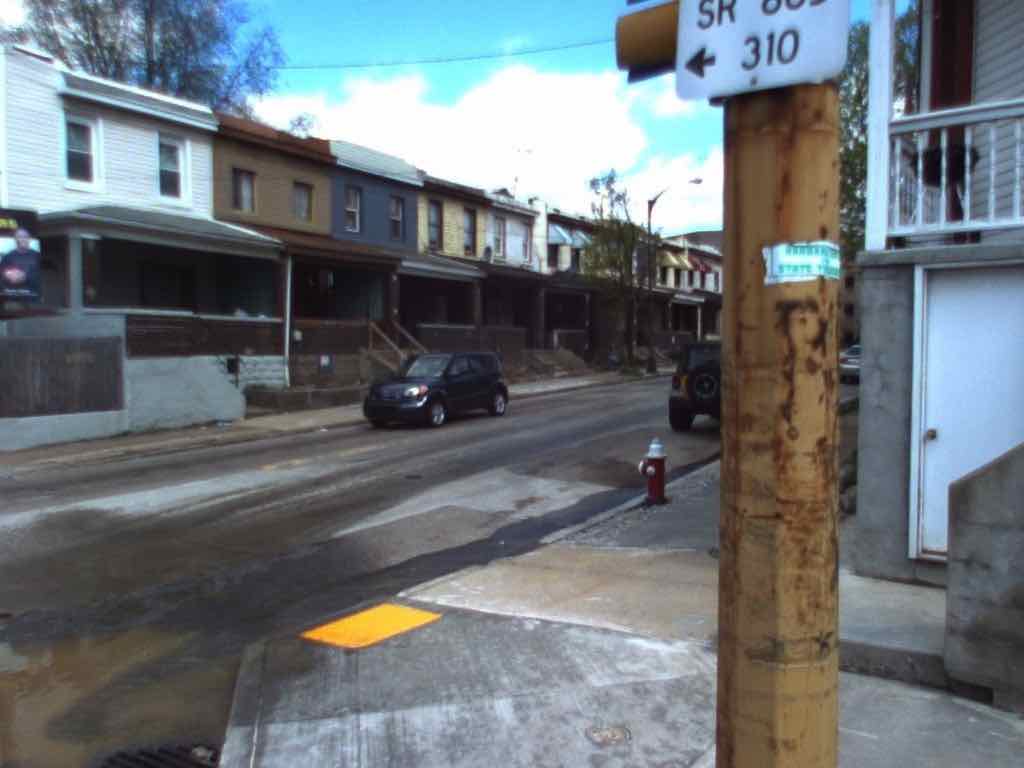}\hfill
	\includegraphics[width=0.19\textwidth]{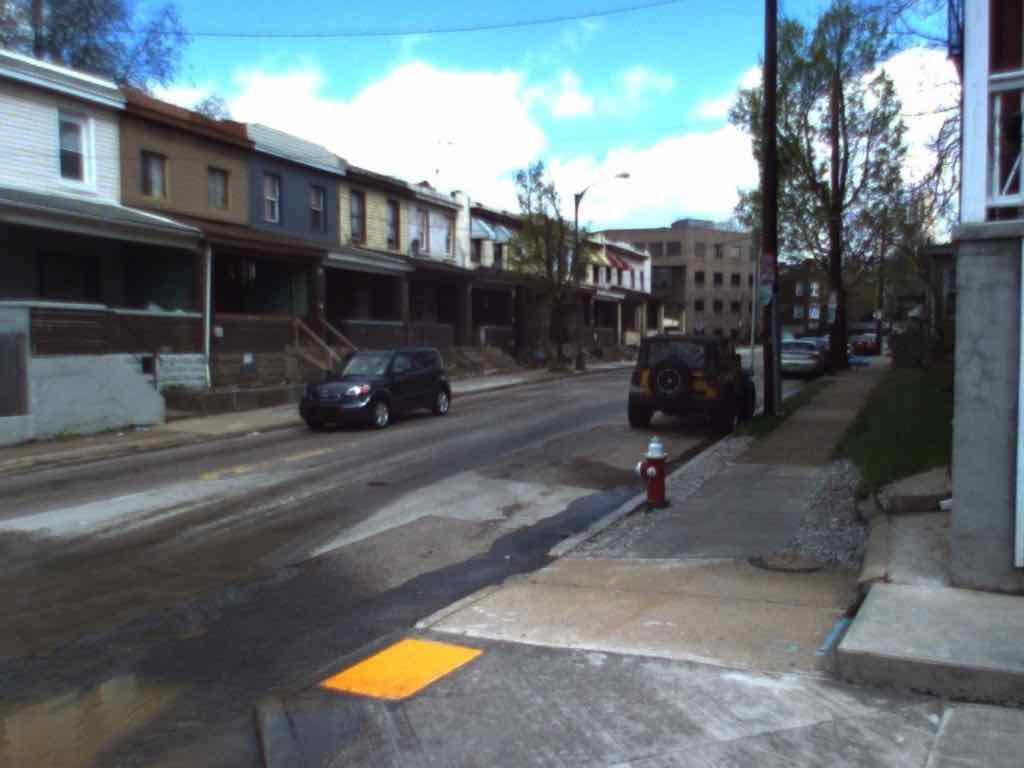}\hfill
	\includegraphics[width=0.19\textwidth]{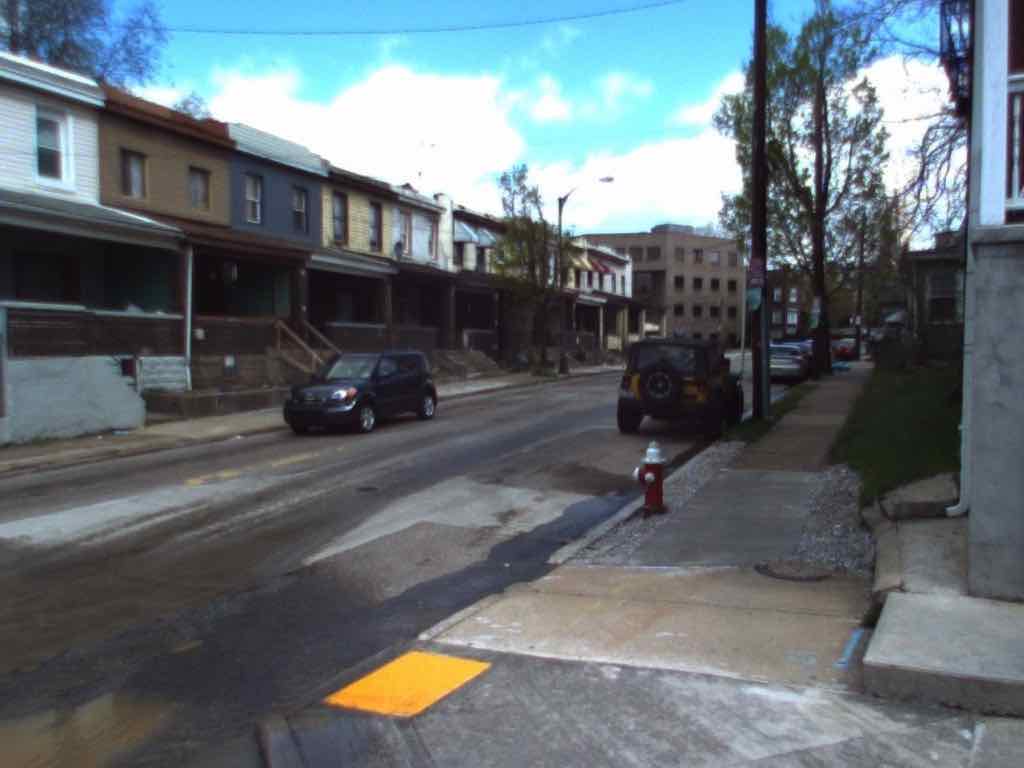}\hfill
	\includegraphics[width=0.19\textwidth]{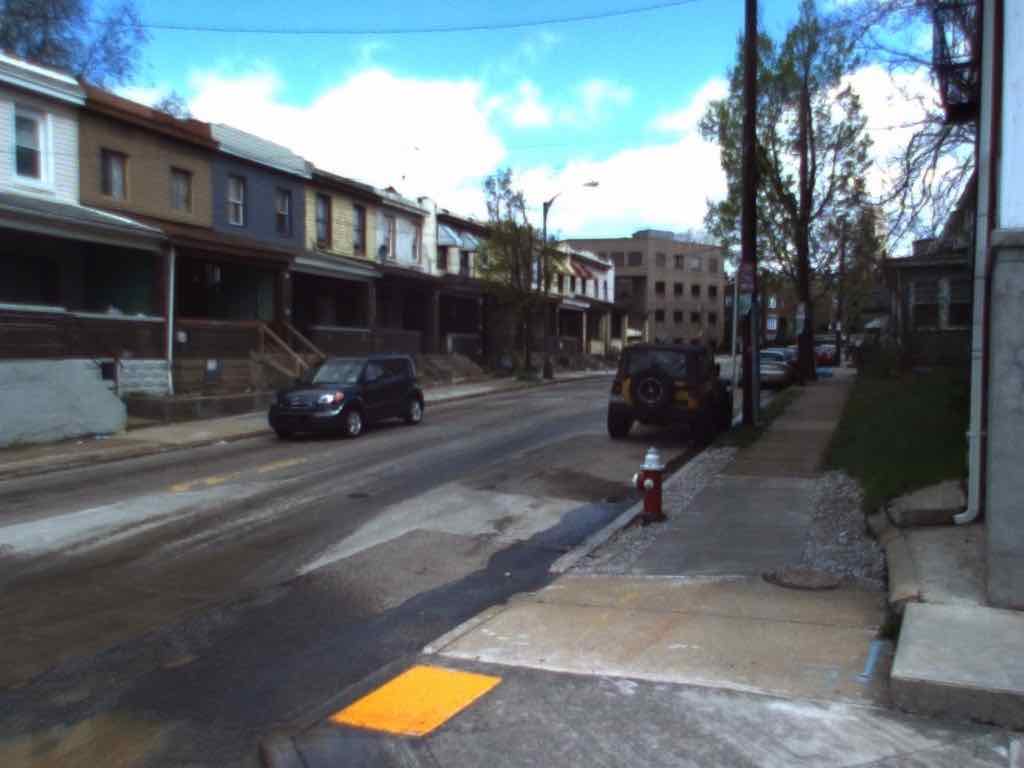}\hfill 
	\caption{A sample sequence from ``Street Change.'' Ground Truth Annotations: ``Construction signs are gone from the street,'' ``the road sign is gone,'' ``the construction sign is gone,'' ``the construction signs are gone.'' \textcolor{red}{Generated Annotations}: \textcolor{red}{``the street sign is missing,'' ``the construction work is done,'' ``the signs were placed.''}
		Note the visual distortions and occlusion in the initial half of the sequence, which is present in several other sequences in the dataset.}\label{fig:vlcmu-1}
\end{figure}

\subsection{Example Street Change Sequences}

Here we illustrate some sample full sequences from the Street Change Dataset, along with the associated ground-truth annotations, and the annotations produced by our method with 100 labeled training sequences and 30 unlabeled training sequences. The displayed images are drawn from the test sequences.  To generate multiple captions, different pairs in the sequence were sampled; the language model is sampled greedily for all pairs.

All images best viewed digitally.

\begin{figure}
	\includegraphics[width=0.16\textwidth]{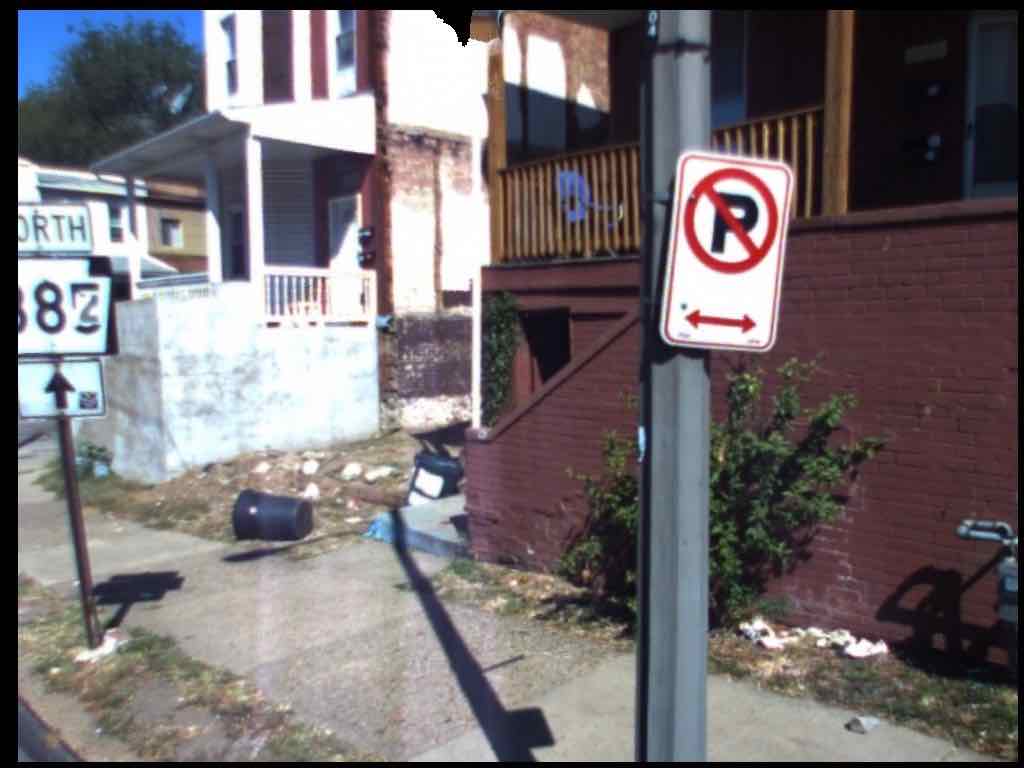}\hfill
	\includegraphics[width=0.16\textwidth]{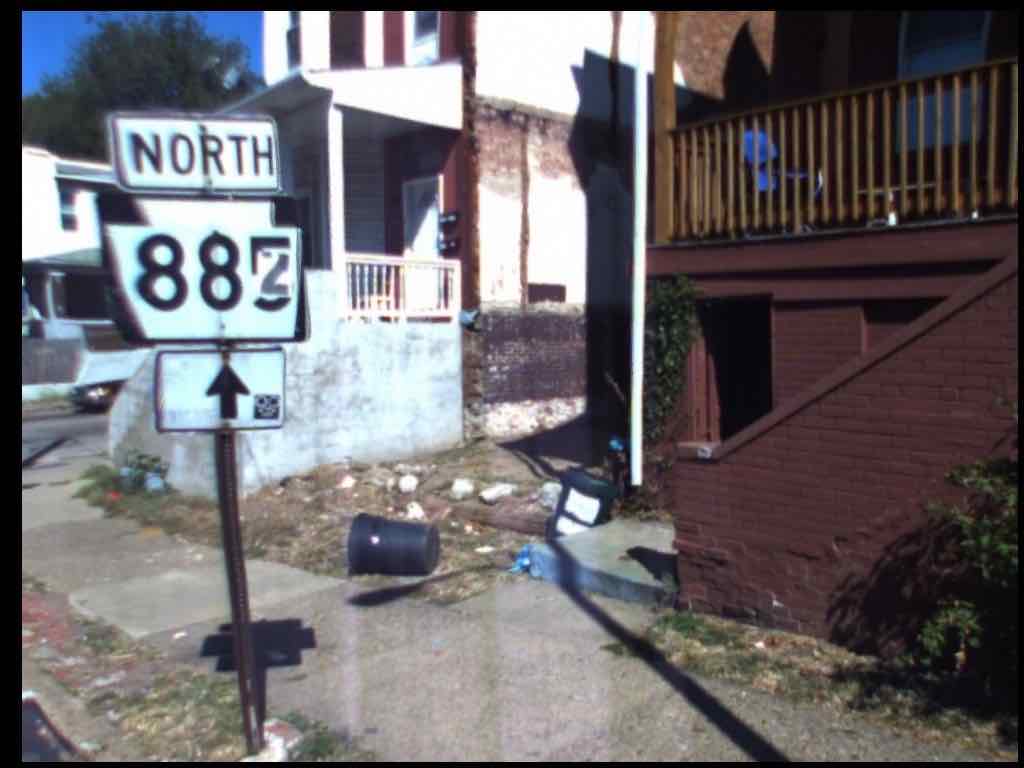}\hfill
	\includegraphics[width=0.16\textwidth]{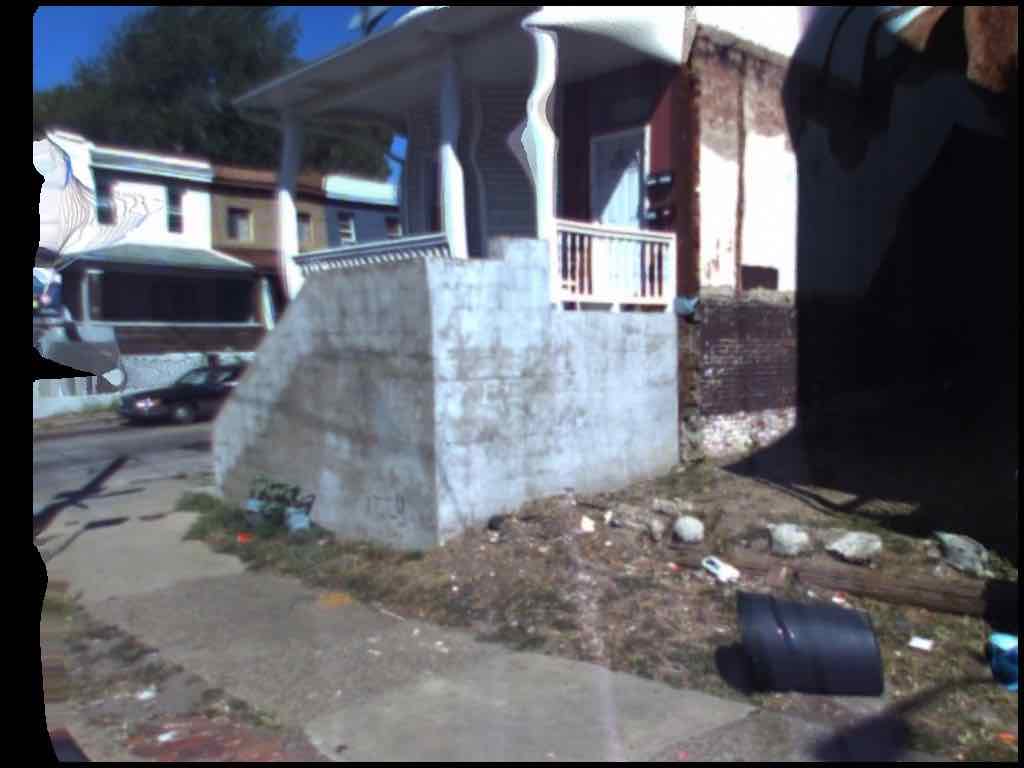}\hfill 
	\includegraphics[width=0.16\textwidth]{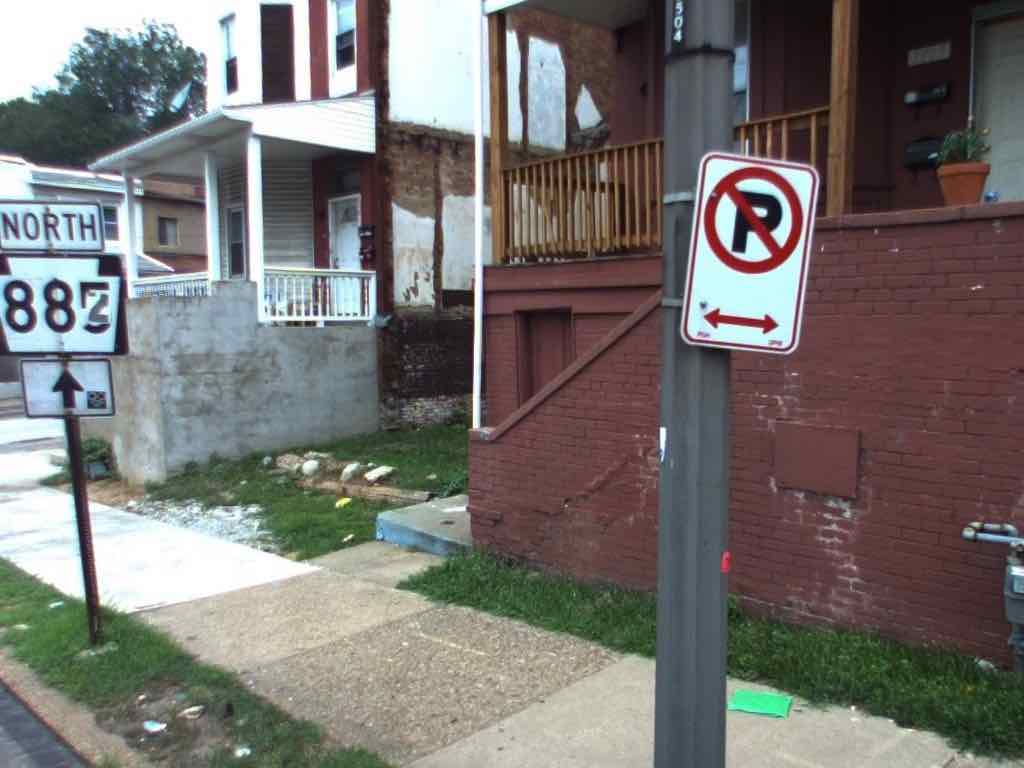}\hfill
	\includegraphics[width=0.16\textwidth]{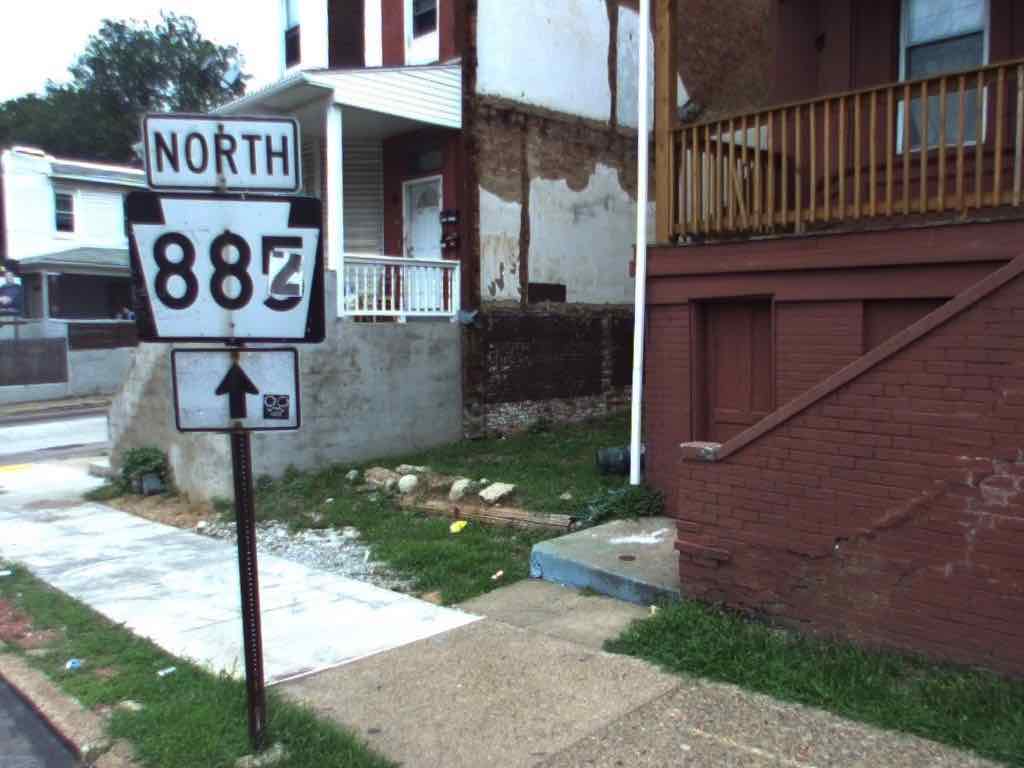}\hfill 
	\includegraphics[width=0.16\textwidth]{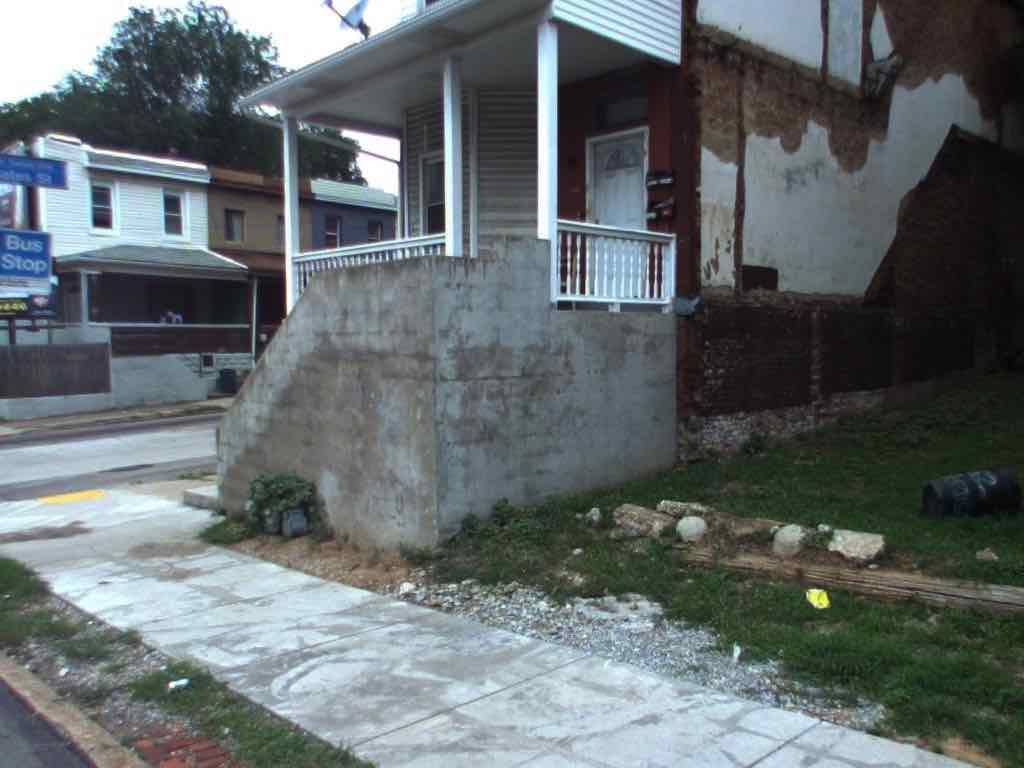}\hfill
	\caption{A sample sequence from ``Street Change.'' Ground Truth Annotations: ``The trash is gone,'' ``the garbage can is gone,'' ``the bush is no longer there,'' ``the garbage can has been removed,'' ``the yard now has grass.'' \textcolor{red}{Generated Annotations}: \textcolor{red}{``The garbage can was removed,'' ``the trash can is gone,'' ``garbage is gone.''}   Shorter sequences tend to contain larger viewpoint changes between frames.}\label{fig:vlcmu-3}
\end{figure}

\begin{figure}
	\includegraphics[width=0.19\textwidth]{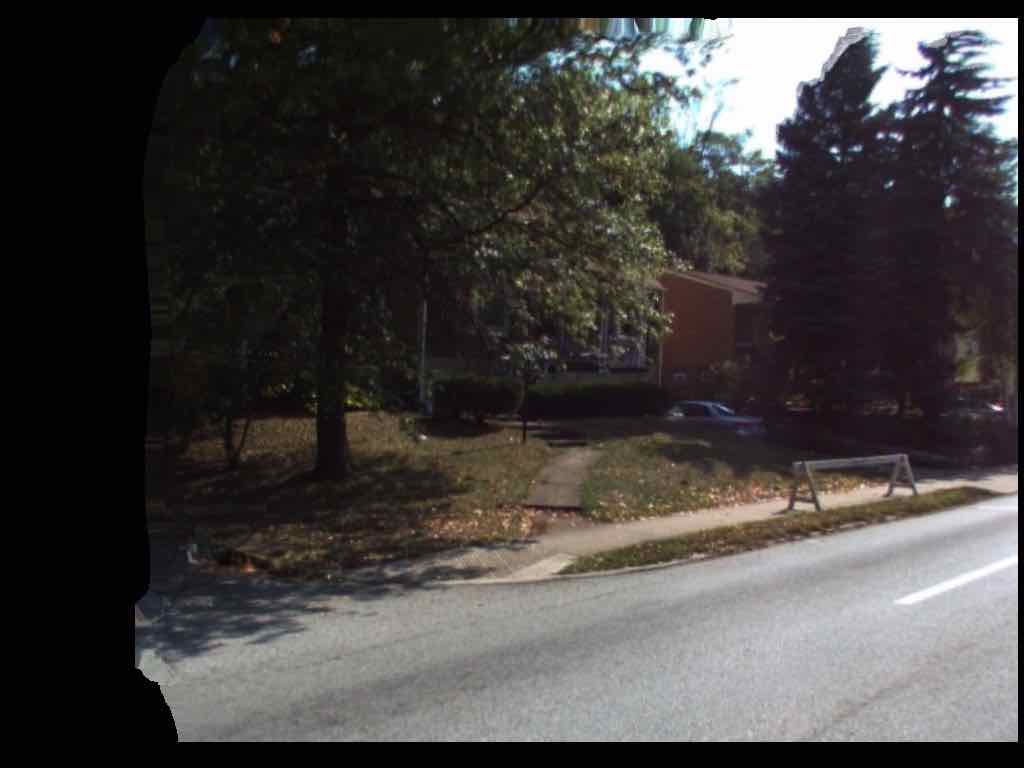}\hfill
	\includegraphics[width=0.19\textwidth]{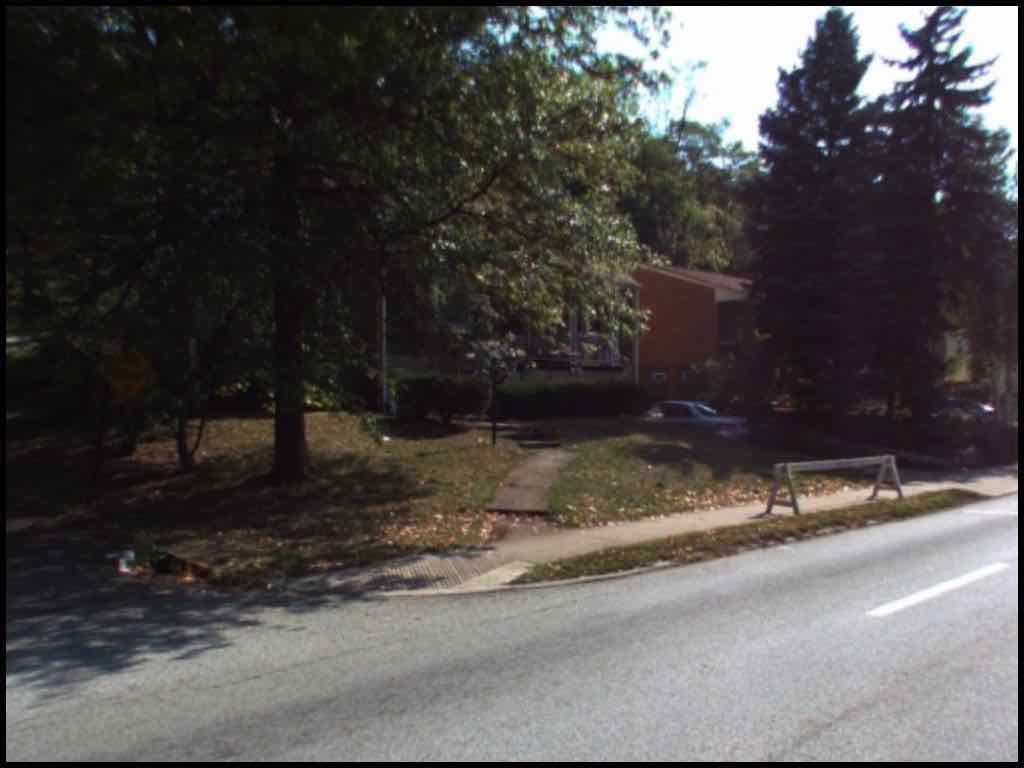}\hfill
	\includegraphics[width=0.19\textwidth]{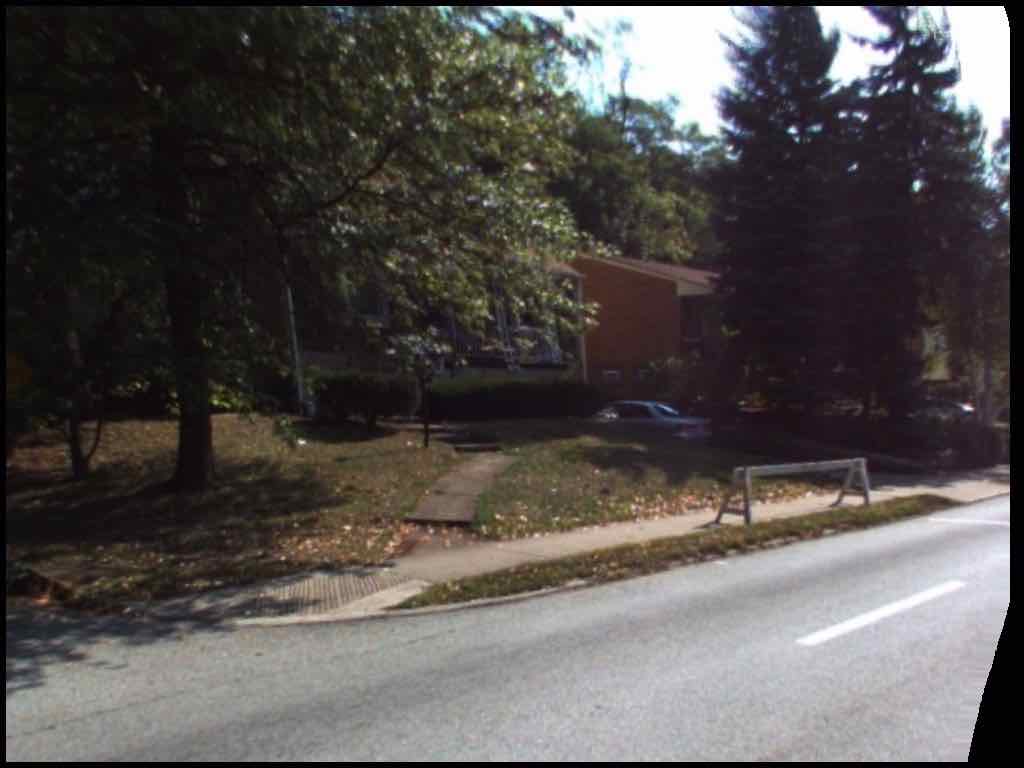}\hfill
	\includegraphics[width=0.19\textwidth]{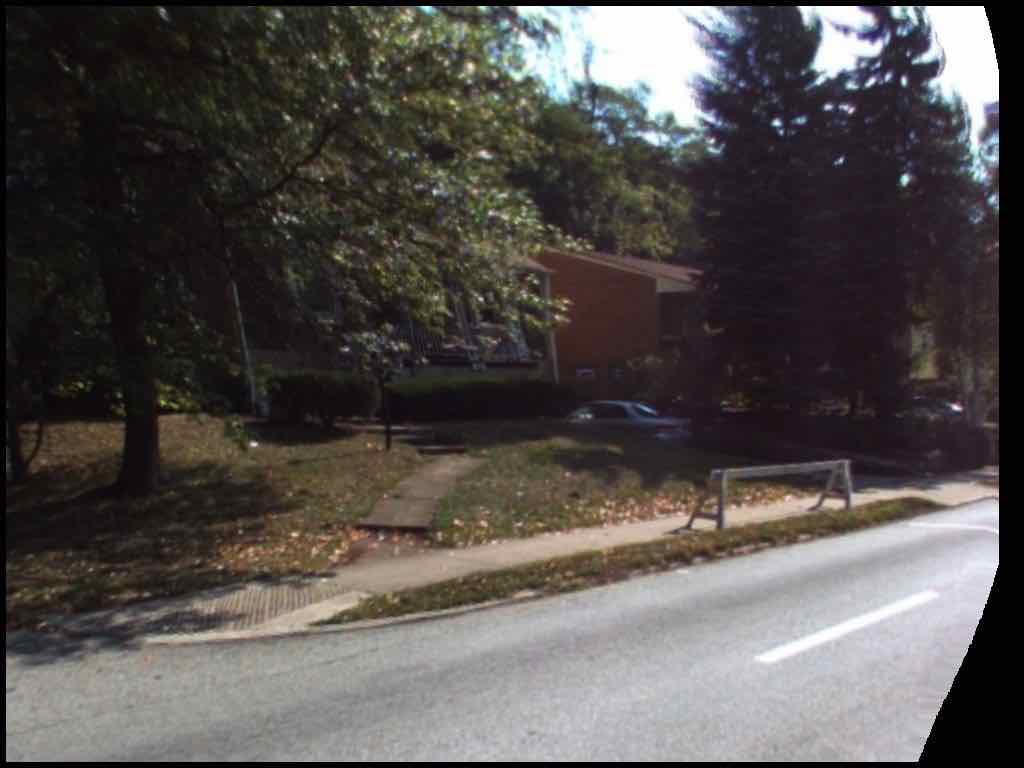}\hfill
	\includegraphics[width=0.19\textwidth]{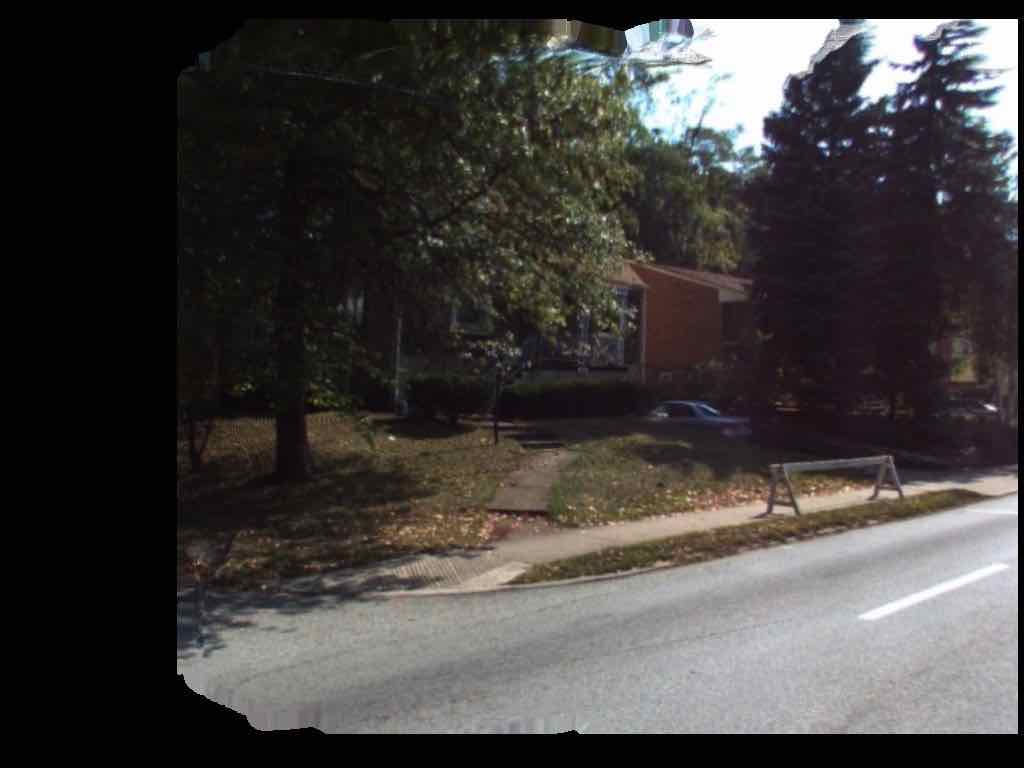}\hfill \\
	\includegraphics[width=0.19\textwidth]{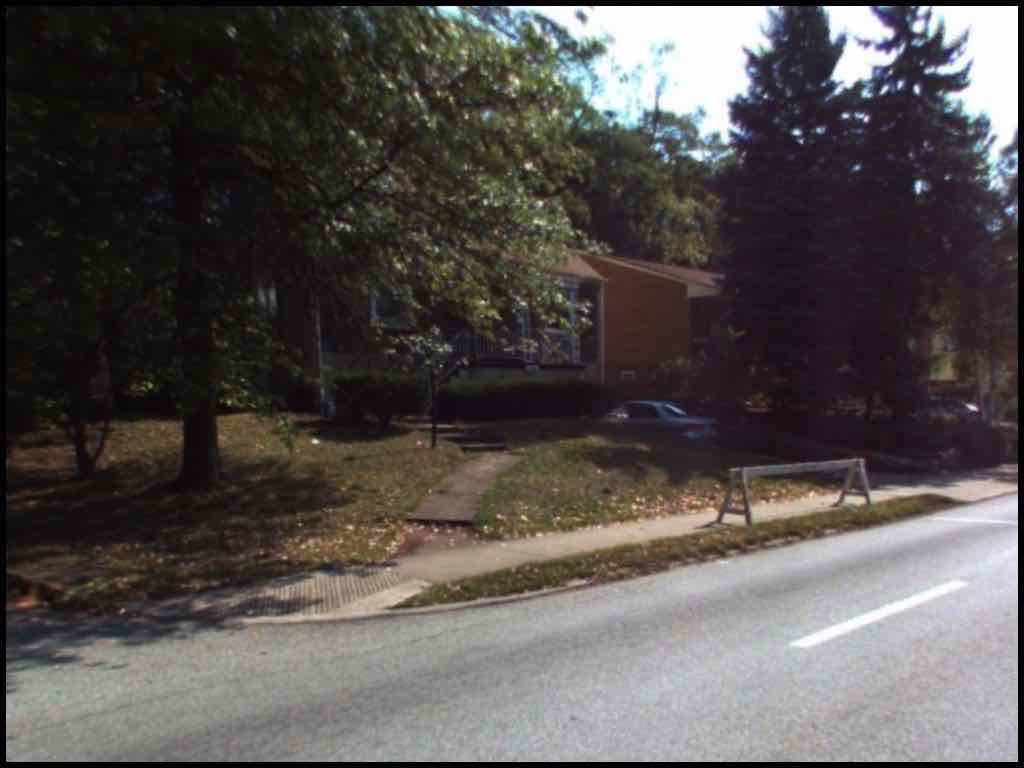}\hfill
	\includegraphics[width=0.19\textwidth]{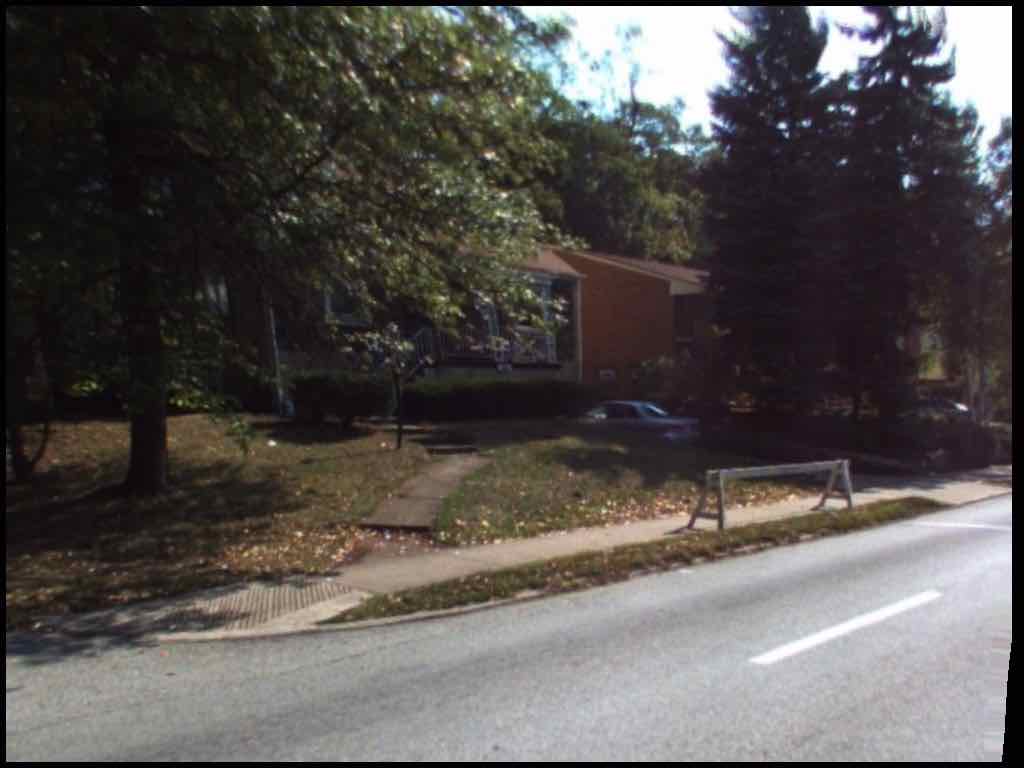}\hfill
	\includegraphics[width=0.19\textwidth]{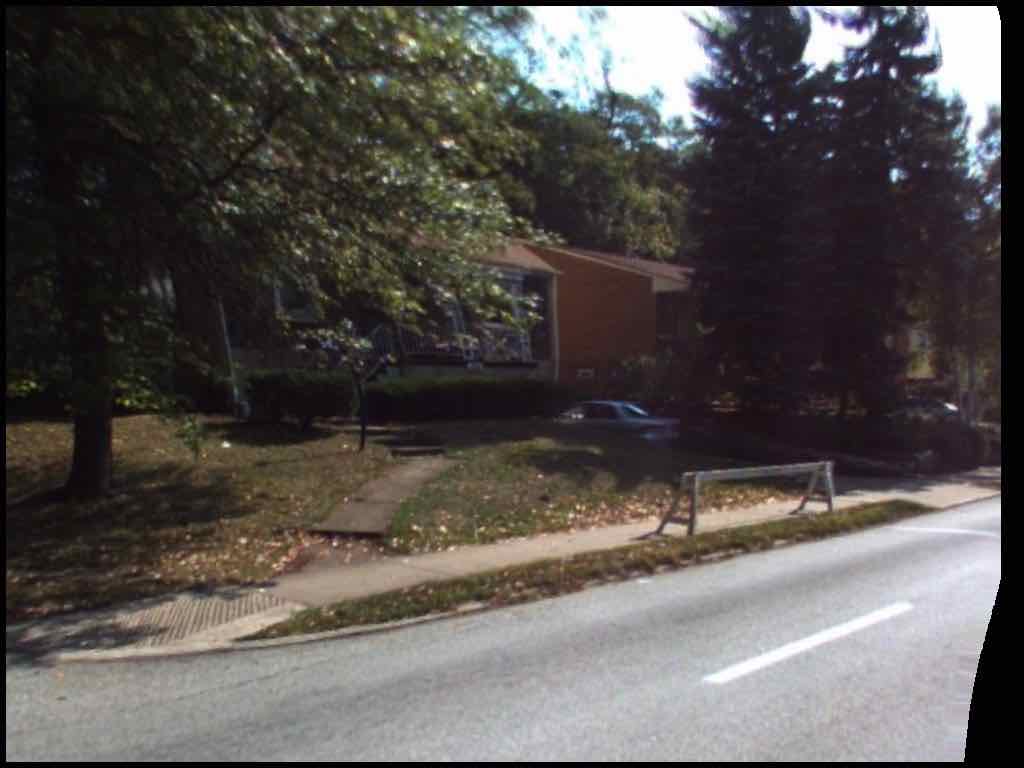}\hfill
	\includegraphics[width=0.19\textwidth]{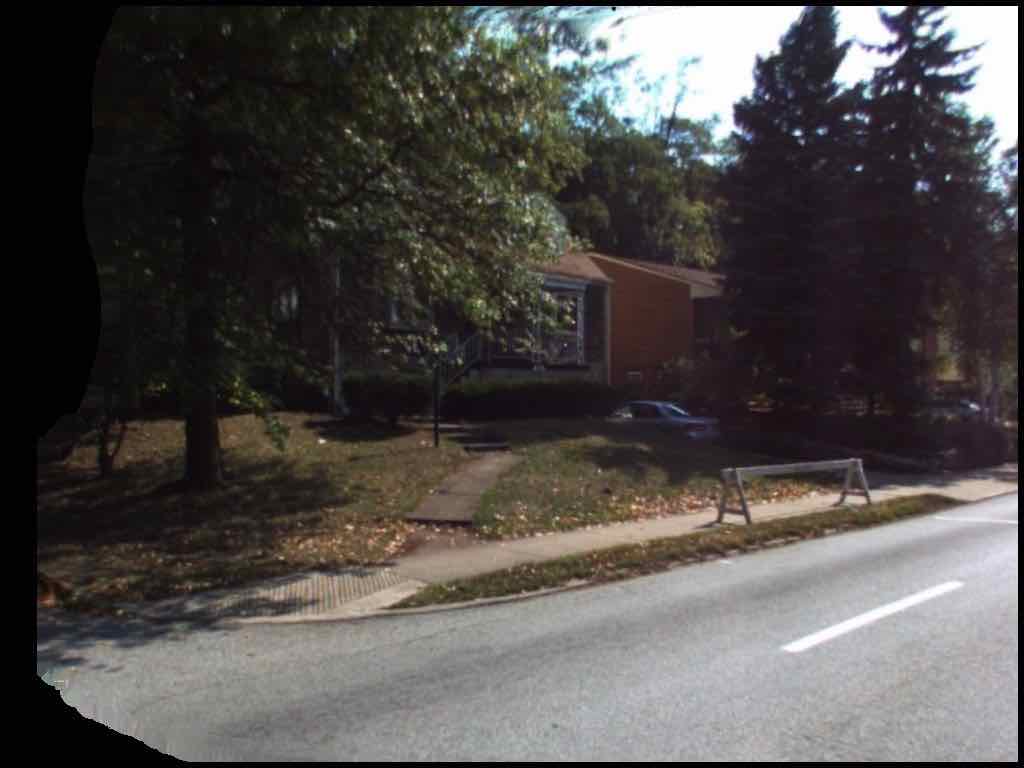}\hfill
	\includegraphics[width=0.19\textwidth]{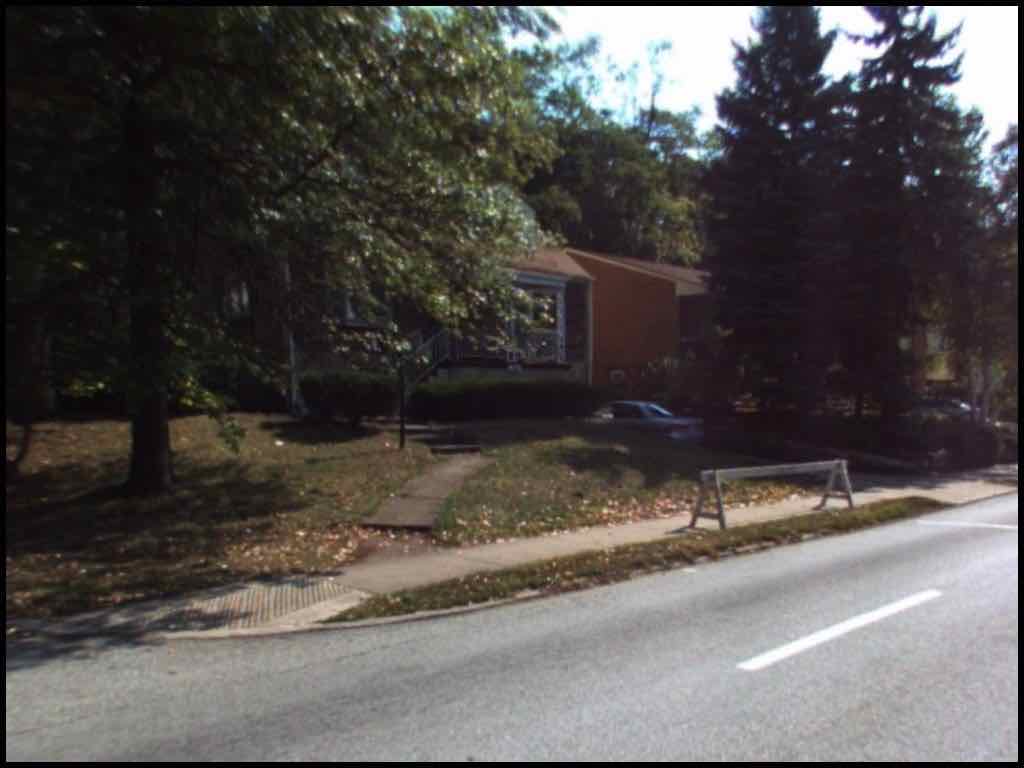}\hfill \\
	\includegraphics[width=0.19\textwidth]{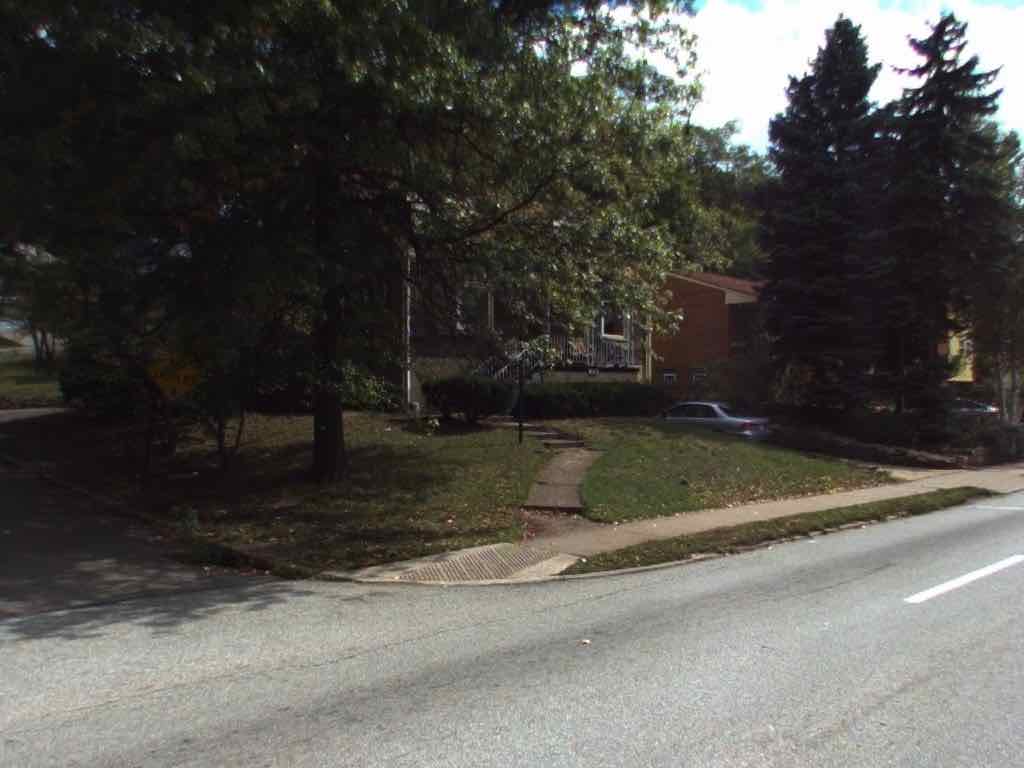}\hfill
	\includegraphics[width=0.19\textwidth]{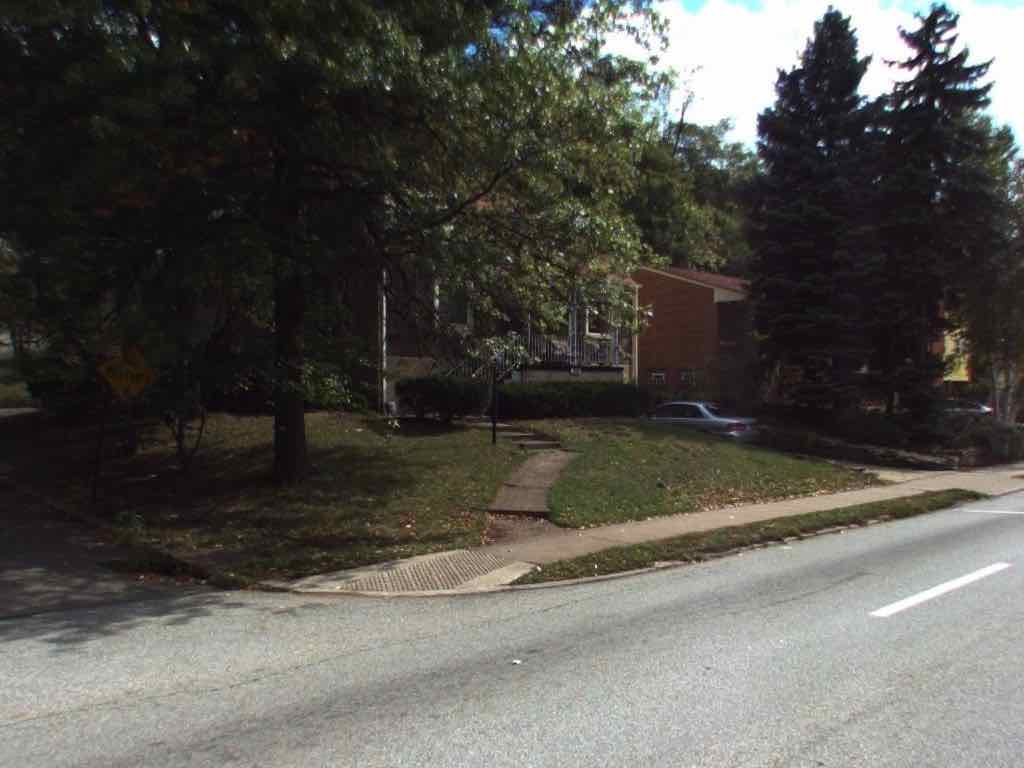}\hfill
	\includegraphics[width=0.19\textwidth]{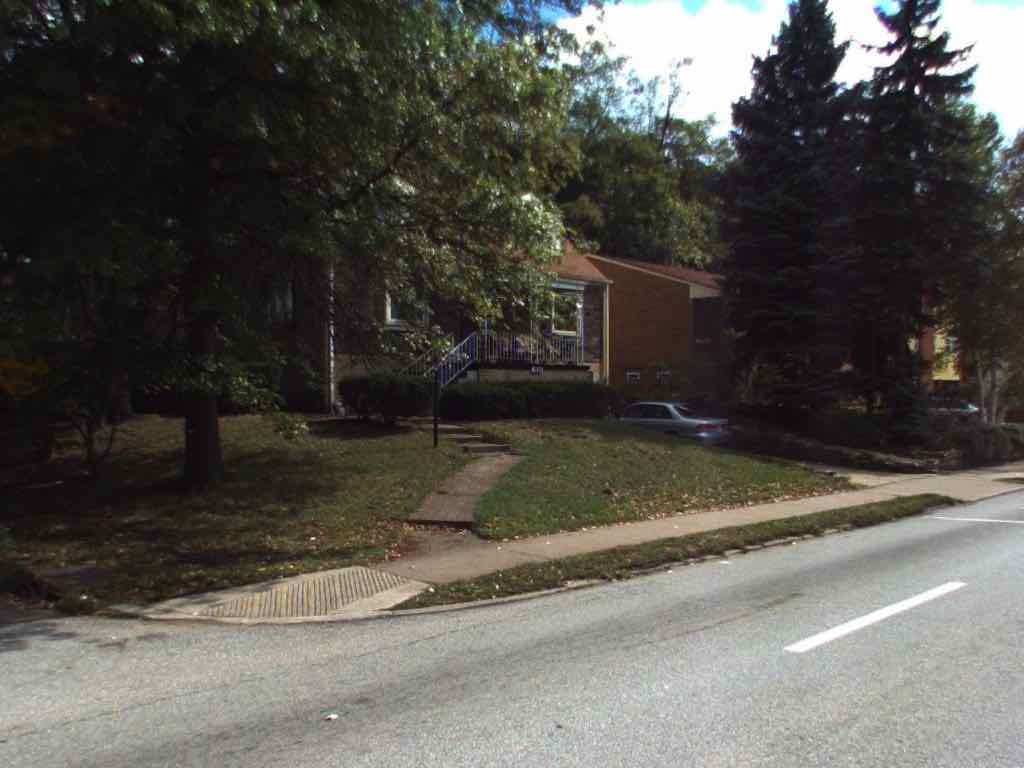}\hfill
	\includegraphics[width=0.19\textwidth]{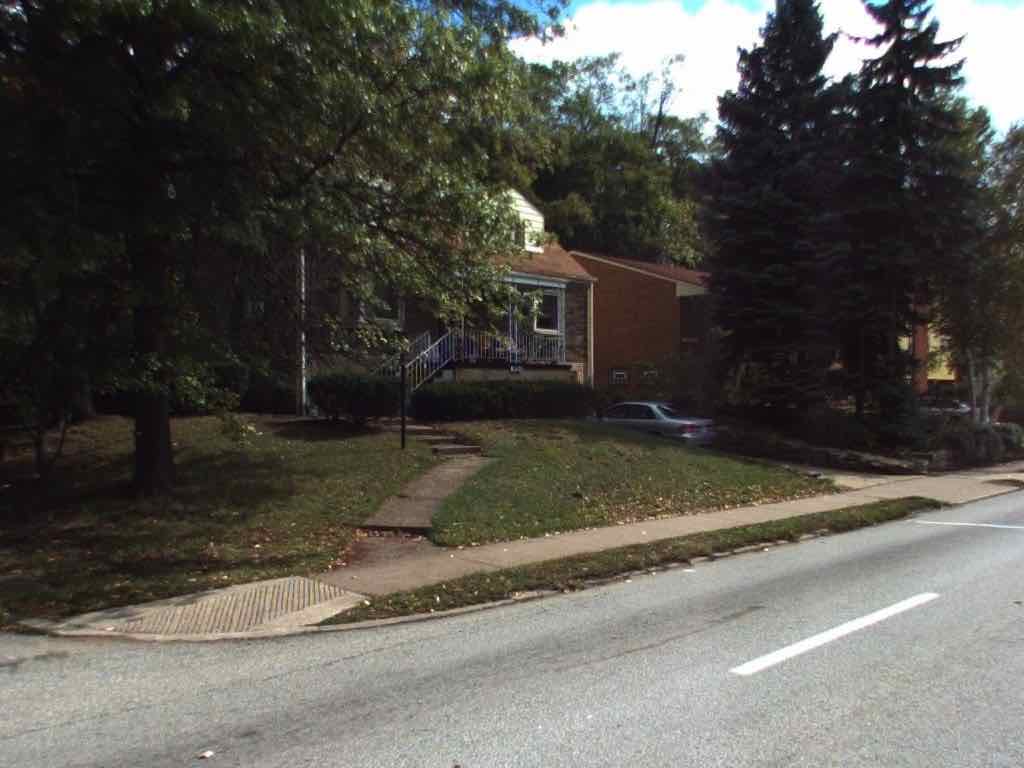}\hfill
	\includegraphics[width=0.19\textwidth]{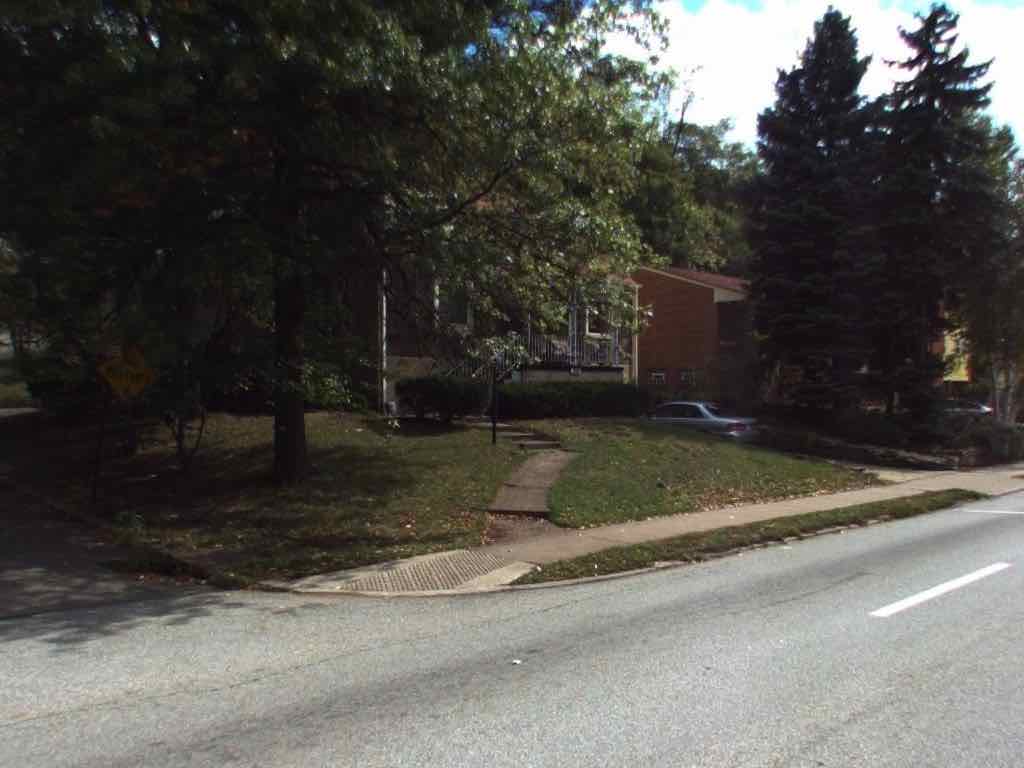}\hfill \\
	\includegraphics[width=0.19\textwidth]{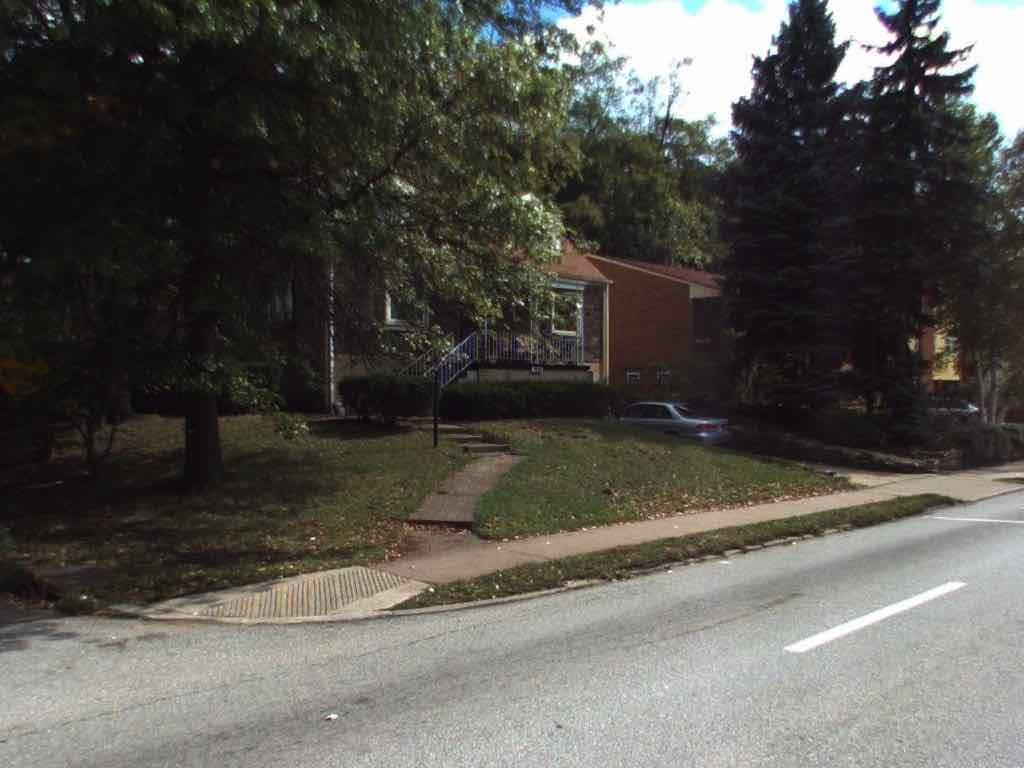}\hfill
	\includegraphics[width=0.19\textwidth]{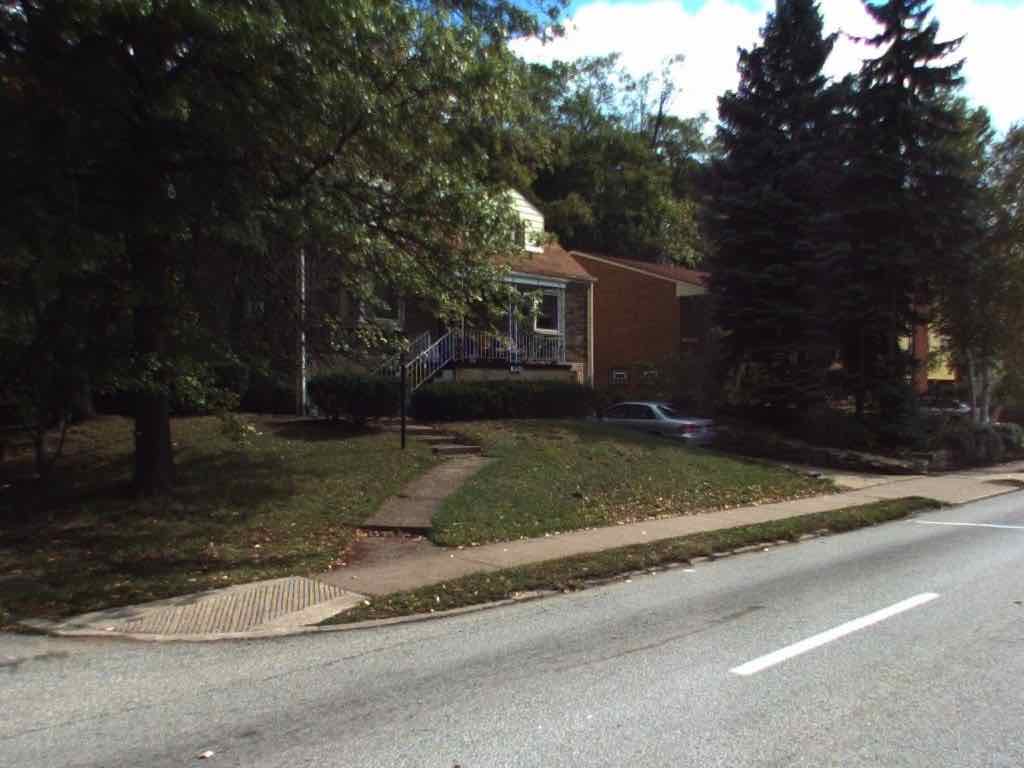}\hfill
	\includegraphics[width=0.19\textwidth]{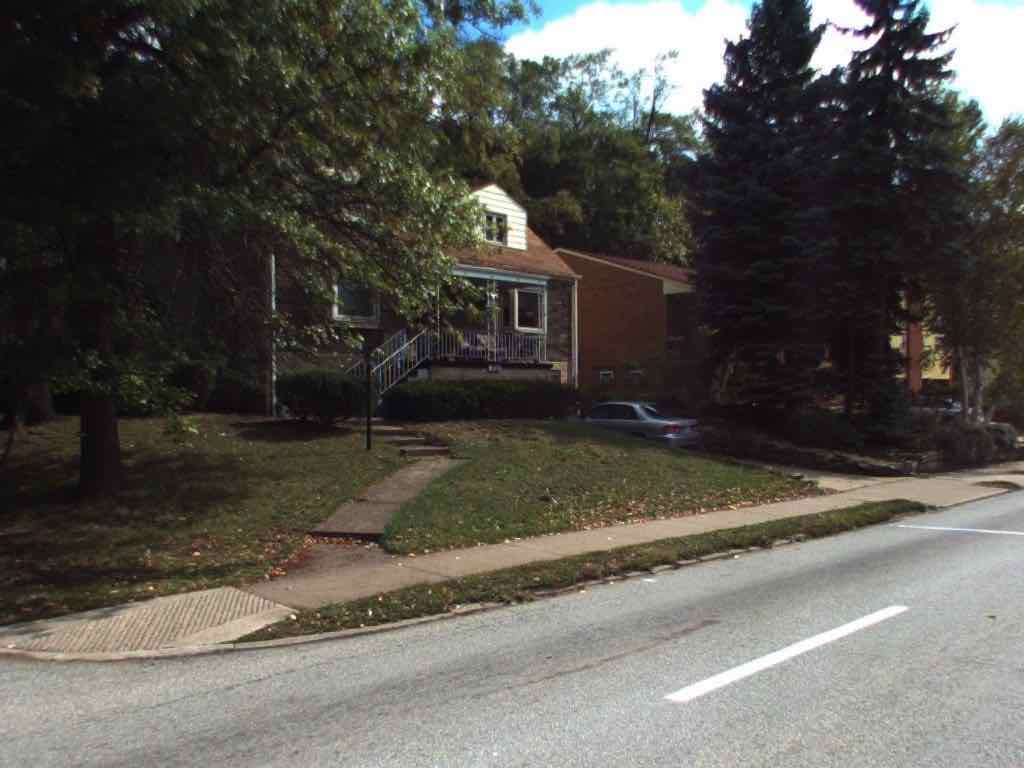}\hfill
	\includegraphics[width=0.19\textwidth]{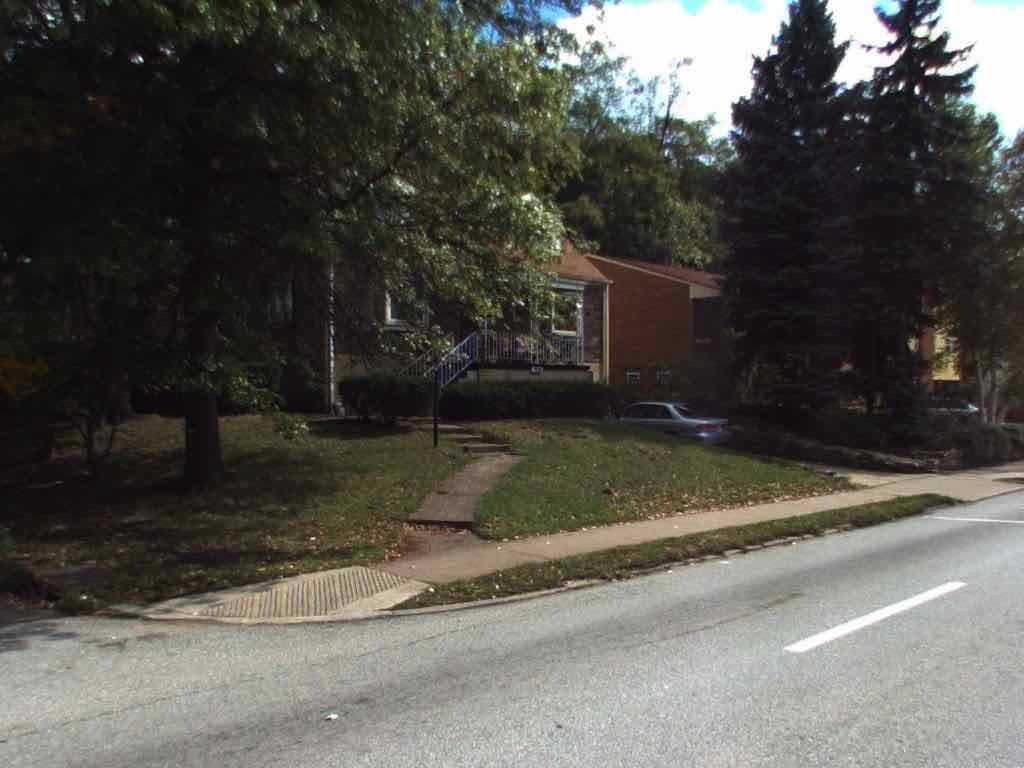}\hfill
	\includegraphics[width=0.19\textwidth]{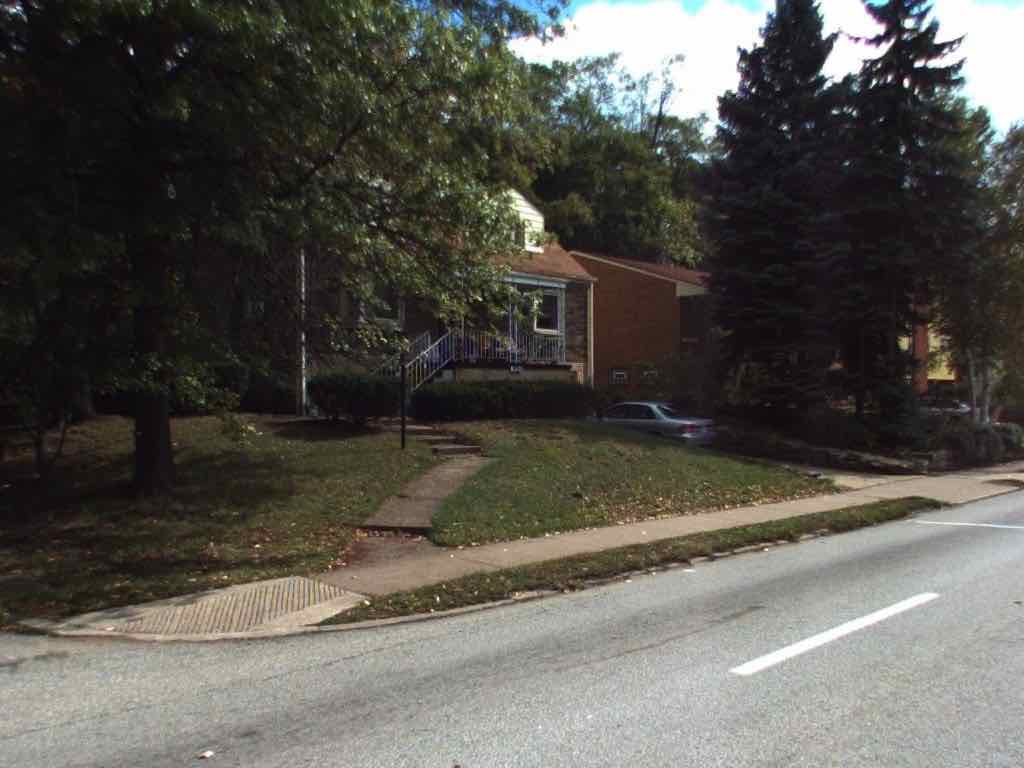}\hfill 
	\caption{A sample sequence from ``Street Change.'' Ground Truth Annotations: ``There is no more sign on sidewalk,'' ``the wooden barrier is gone,'' ``the barricade is gone,'' ``the wooden barricade on the sidewalk disappeared,'' ``the construction barrier is gone,'' ``there is no longer a wooden barrier on the sidewalk.'' \textcolor{red}{Generated Annotations}: \textcolor{red}{``the saw horse is gone,'' ``the construction barricade is gone,'' ``the construction barrier on the sidewalk is no longer there.''}  While visually finding the changepoint is straightforward in some sequences, there are many examples like this sequence, where the visual distinction between the first and second halves of the sequence is subtle.}\label{fig:vlcmu-2}
\end{figure}

\end{document}